\documentclass[sigconf]{acmart}

\usepackage{textcomp}
\usepackage{xcolor}

\AtBeginDocument{%
  \providecommand\BibTeX{{%
    \normalfont B\kern-0.5em{\scshape i\kern-0.25em b}\kern-0.8em\TeX}}}

\usepackage{amsmath}
\usepackage{rotating}
\usepackage{booktabs}
\usepackage{listings}
\usepackage{mathtools}

\usepackage{hyperref}

\DeclarePairedDelimiter\abs{\lvert}{\rvert}%
\DeclarePairedDelimiter\norm{\lVert}{\rVert}%

\newtheorem{theorem}{Theorem}[section]

\newtheorem{lemma}[theorem]{Lemma}
\newtheorem{Assumption}[theorem]{Assumption}

\acmConference[KDD '22]{KDD '22}{August 14--18, 2021}{Virtual Event}
\usepackage{caption}
\usepackage{subcaption}

\usepackage[ruled,linesnumbered]{algorithm2e} 

\newcommand{\hide}[1]{}

\makeatletter
\newcommand*{\rom}[1]{\expandafter\@slowromancap\romannumeral #1@}
\makeatother
\newcommand{\matr}[1]{\boldsymbol{#1}}
\newcommand{\vect}[1]{\boldsymbol{#1}}

\newcommand{\name}{{\textsc{AGG-UCB}}}

\DeclarePairedDelimiterX{\inp}[2]{\langle}{\rangle}{#1, #2}


\copyrightyear{2022} 
\acmYear{2022} 
\setcopyright{acmcopyright}\acmConference[KDD '22]{Proceedings of the 28th ACM SIGKDD Conference on Knowledge Discovery and Data Mining}{August 14--18, 2022}{Washington, DC, USA}
\acmBooktitle{Proceedings of the 28th ACM SIGKDD Conference on Knowledge Discovery and Data Mining (KDD '22), August 14--18, 2022, Washington, DC, USA}
\acmPrice{15.00}
\acmDOI{10.1145/3534678.3539312}
\acmISBN{978-1-4503-9385-0/22/08}

\begin{document}

\title{Neural Bandit with Arm Group Graph}

\author{Yunzhe Qi}
\affiliation{%
  \institution{University of Illinois at Urbana-Champaign}
  \country{}
 }
\email{yunzheq2@illinois.edu}

\author{Yikun Ban}
\affiliation{%
  \institution{University of Illinois at Urbana-Champaign}
  \country{}
} 
\email{yikunb2@illinois.edu}

\author{Jingrui He}
\affiliation{%
  \institution{University of Illinois at Urbana-Champaign}
  \country{}
}
\email{jingrui@illinois.edu}

\begin{abstract}
    Contextual bandits aim to identify among a set of arms the optimal one with the highest reward based on their contextual information. Motivated by the fact that the arms usually exhibit group behaviors and the mutual impacts exist among groups, we introduce a new model, Arm Group Graph (AGG),  where the nodes represent the groups of arms and the weighted edges formulate the correlations among groups. To leverage the rich information in AGG, we propose a bandit algorithm, AGG-UCB, where the neural networks are designed to estimate rewards, and we propose to utilize graph neural networks (GNN) to learn the representations of arm groups with correlations. To solve the exploitation-exploration dilemma in bandits, we derive a new upper confidence bound (UCB) built on neural networks (exploitation) for exploration. Furthermore, we prove that AGG-UCB can achieve a near-optimal regret bound with over-parameterized neural networks, and provide the convergence analysis of GNN with fully-connected layers which may be of independent interest. In the end, we conduct extensive experiments against state-of-the-art baselines on multiple public data sets, showing the effectiveness of the proposed algorithm. 
   
\end{abstract}


\maketitle

\section{Introduction} \label{sec_introduction}
Contextual bandits are a specific type of multi-armed bandit (MAB) problem where the learner has access to the contextual information (contexts) related to arms at each round, and the learner is required to make recommendations based on past contexts and received rewards.
A variety of models and algorithms have been proposed and successfully applied on real-world problems, such as online content and advertising recommendation \cite{linucb-Ind_li2010contextual,colla_environ_2016}, clinical trials \cite{MAB_clinical_trial-durand2018contextual,MAB_clinical_trial_2_villar2015multi} and virtual support agents \cite{virtual_support_bot_sajeev2021contextual}.

In this paper, we focus on exploiting the accessible arm information to improve the performance of bandit algorithms. 
Among different types of contextual bandit algorithms, upper confidence bound (UCB) algorithms have been proposed to balance between exploitation and exploration \cite{UCB_auer2002finite,Lin_UCB-2011,kernel_ucb-2013}. For conventional UCB algorithms, they are either under the "pooling setting" \cite{Lin_UCB-2011}
where one single UCB model is applied for all candidate arms, or the "disjoint setting" \cite{linucb-Ind_li2010contextual} where each arm is given its own estimator without the collaboration across different arms. Both settings have their limitations: applying only one single model may lead to unanticipated estimation error when some arms exhibit distinct behaviors \cite{dynamic_ensemble_wu2019dynamic,colla_environ_2016}; on the other hand, assigning each arm its own estimator neglects the mutual impacts among arms and usually suffers from limited user feedback \cite{CAB_2017,local_clustering-ban2021local}.

To deal with this challenge, adaptively assigning UCB models to arms based on their group information can be an ideal strategy, i.e., each group of arms has one estimator to represent its behavior. This modeling strategy is linked to "arm groups" existing in real-world applications.
For example, regarding the online movie recommendation scenario, the movies (arms) with the same genre can be assigned to one (arm) group. 
Another scenario is the drug development, where given a new cancer treatment and a patient pool, we need to select the best patient on whom the treatment is most effective. Here, the patients are the arms, and they can be naturally grouped by their non-numerical attributes, such as the cancer types.
Such group information is easily accessible, and can significantly improve the performance of bandit algorithms. 
%
Although some works ~\cite{co_filter_bandits_2016,kmtl-ucb_2017} have been proposed to leverage the arm correlations, they can suffer from two common limitations. First, they rely on the assumption of parametric (linear / kernel-based) reward functions, which may not hold in real-world applications \cite{Neural-UCB}. Second, they both neglect the correlations among arm groups.
We emphasize that the correlations among arm groups also play indispensable roles in many decision-making scenarios.
For instance, in online movie recommendation, with each genre being a group of movies, the users who like "adventure movies" may also appreciate "action movies". Regarding drug development, since the alternation of some genes may lead to multiple kinds of tumors \cite{correlated_cancer-shao2019copy}, different types of cancer can also be correlated to some extent. 

To address these limitations, we first introduce a novel model, AGG (Arm Group Graph), to formulate non-linear reward assumptions and arm groups with correlations. In this model, as arm attributes are easily accessible (e.g., movie's genres and patient's cancer types), the arms with the same attribute are assigned into one group, and represented as one node in the graph.
The weighted edge between two nodes represents the correlation between these two groups. 
In this paper, we assume the arms from the same group are drawn from one unknown distribution. This also provides us with an opportunity to model the correlation of two arm groups by modeling the statistical distance between their associated distributions. Meantime, the unknown non-parametric reward mapping function can be either linear or non-linear.

Then, with the arm group graph, we propose the \name\ framework for contextual bandits. It applies graph neural networks (GNNs) to learn the representations of arm groups with correlations, and neural networks to estimate the reward functions (exploitation).
In particular, with the collaboration across arm groups, each arm will be assigned with the group-aware arm representation learned by GNN, which will be fed into a fully-connected (FC) network for the estimation of arm rewards.
To deal with the exploitation-exploration dilemma, we also derive a new upper confidence bound based on network-gradients for exploration.
By leveraging the arm group information and modeling arm group correlations, our proposed framework provides a novel arm selection strategy for dealing with the aforementioned challenges and limitations.
Our main contributions can be summarized as follows: 
\begin{itemize}
    \item First, motivated by real-world applications, we introduce a new graph-based model in contextual bandits to leverage the available group information of arms and exploit the potential correlations among arm groups.
    \item Second, we propose a novel UCB-based neural framework called \name~ for the graph-based model. To exploit the relationship of arm groups, \name~ estimates the arm group graph with received contexts on the fly, and utilizes GNN to learn group-aware arm representations.
    \item Third,  we prove that AGG-UCB can achieve a near-optimal regret bound in the over-parameterized neural works, and provide convergence analysis of GNN with fully-connected layers, which may be of independent interest.
    \item Finally, we conduct experiments on publicly available real data sets, and demonstrate that our framework outperforms state-of-the-art techniques. Additional studies are conducted to understand the properties of the proposed framework.
\end{itemize}

The rest of this paper is organized as following. In Section \ref{sec_related_works}, we briefly discuss related works. Section \ref{sec_problem_def} introduces the new problem settings, and details of our proposed framework \name~ will be presented in Section \ref{sec_proposed_framework}. Then, we provide theoretical analysis for \name~in Section \ref{sec_theoretical_analysis}. After presenting experimental results in Section \ref{sec_experiments}, we finally conclude the paper in Section \ref{sec_conclusion}. Due to the page limit, readers may refer to our arXiv version of the paper for the supplementary contents 
(\url{https://arxiv.org/abs/2206.03644}).

\section{Related Works} \label{sec_related_works}
In this section, we briefly review the related work on contextual bandits.
Lin-UCB \cite{Lin_UCB-2011} first formulates the reward estimation through a linear regression with the received context and builds a confidence bound accordingly. Kernel-UCB \cite{kernel_ucb-2013} further extends the reward mapping to the Reproducing Kernel Hilbert Space (RKHS) for the reward and confidence bound estimation under non-linear settings. Besides, there are algorithms under the non-linear settings. Similarly, CGP-UCB \cite{CGP_krause2011contextual} models the reward function through a Gaussian process. GCN-UCB \cite{GCN_UCB-upadhyay2020online} applies the GNN model to learn each context an embedding for the linear regression.
Then, Neural-UCB \cite{Neural-UCB} proposes to apply FC neural network for reward estimations and derive a confidence bound with the network gradient, which is proved to be a success, and similar ideas has been applied to some other models \cite{neural_multifacet-ban2021multi,neural_thompson-zhang2020neural,CNN_UCB-ban2021convolutional}. 
\cite{ban2022eenet} assigns another FC neural network to learn the confidence ellipsoid for exploration.
Yet, as these works consider no collaboration among estimators, they may suffer from the performance bottleneck in the introduction. 

To collaborate with different estimators for contextual bandits, various approaches are proposed from different perspectives. User clustering algorithms \cite{club_2014,SCLUB_li2019improved,local_clustering-ban2021local,Ban2022NeuralCF} try to cluster user with alike preferences into user groups for information sharing while COFIBA \cite{co_filter_bandits_2016} additionally models the relationship of arms. 
Then, KMTL-UCB \cite{kmtl-ucb_2017} extends Kernel-UCB to multi-task learning settings for a refined reward and confidence bound estimation. 
However, these works may encounter with performance bottlenecks as they incline to make additional assumptions on the reward mapping by applying parametric models and neglect the available arm group information.

GNNs \cite{GCN_kipf2016semi,SGC_wu2019simplifying,APPNP-klicpera2018predict,SDG_fu2021sdg} are a kind of neural models that deal with tasks on graphs, such as community detection \cite{GNN_community_detection-you2019position}, recommender systems \cite{GNN_Rec_wei2020fast} and modeling protein interactions \cite{DPPIN_fu2021dppin}. 
GNNs can learn from the topological graph structure information and the input graph signal simultaneously, which enables \name~ to cooperate with different arm groups by sharing information over the arm group neighborhood.


\section{Problem Definition and Notation} \label{sec_problem_def}
We suppose a fixed pool $\mathcal{C} = \{1, \dots, N_{c}\}$ for arm groups with the number of arm groups being $\abs{\mathcal{C}} = N_{c}$, and assume each arm group $c \in \mathcal{C}$ is associated with an unknown fixed context distribution $\mathcal{D}_{c}$. At each time step $t$, we will receive a subset of groups $\mathcal{C}_{t} \subseteq \mathcal{C}$.
For each group $c \in \mathcal{C}_{t}$, we will have the set of sampled arms $\mathcal{X}_{c, t} = \{\vect{x}^{(1)}_{c, t}, \cdots \vect{x}^{(n_{c, t})}_{c, t}\}$ with the size of $\abs{\mathcal{X}_{c, t}} = n_{c, t}$.
Then, $\forall i \in [n_{c, t}] = \{1, \dots,  n_{c, t}\}$, we suppose $\vect{x}^{(i)}_{c, t} \sim \mathcal{D}_{c}$ with the dimensionality $\vect{x}^{(i)}_{c, t} \in \mathbb{R}^{d_{x}}$. 
Therefore, in the $t$-th round, we receive
\begin{equation}
\{ \mathcal{X}_{c, t} | c \in  \mathcal{C}_{t}\}  \ \text{and} \  \mathcal{X}_{c, t} = \{\vect{x}^{(1)}_{c, t}, \cdots \vect{x}^{(n_{c, t})}_{c, t}\}, \forall c \in  \mathcal{C}_{t}.
\end{equation}
With $\matr{W}^{*} \in \mathbb{R}^{N_{c}\times N_{c}}$ being the unknown affinity matrix encoding the true arm group correlations, the true reward $r_{c, t}^{(i)}$ for arm $\vect{x}^{(i)}_{c, t}$ is defined as
\begin{equation}
\begin{split}
    r_{c, t}^{(i)} = h(\matr{W}^{*}, ~~ \vect{x}^{(i)}_{c, t}) + \epsilon^{(i)}_{c, t}
\end{split}
\label{reward_func_defi}
\end{equation}
where $h(\cdot)$ represents the unknown reward mapping function, 
and $\epsilon$ is the zero-mean Gaussian noise. 
For brevity, let $\vect{x}_{t}$ be the arm we select in round $t$ and $r_t$ be the corresponding received reward.

Our goal is recommending arm $\vect{x}_{t}$ (with reward $r_{t}$) 
at each time step $t$ to minimize the cumulative pseudo-regret $R(T) = \sum_{t=1}^{T} \mathbb{E}\left[  (r_{t}^{*} -  r_{t}) \right]$ where $\mathbb{E}[r_{t}^{*}] = \max_{(c\in\mathcal{C}_{t}, i\in [n_{c, t}])} \big[h(\matr{W}^{*}, \vect{x}^{(i)}_{c, t})\big]$. 
At each time step $t$, the overall set of received contexts is defined as $\mathcal{X}_{t} = \{ \vect{x}^{(i)}_{c, t} \}_{c \in \mathcal{C}_{t}, i\in [n_{c, t}]}$. 
Note that one arm is possibly associated with multiple arm groups, such as a movie with multiple genres. In other words, for some $c, c'\in \mathcal{C}_{t}$, we may have $\mathcal{X}_{c, t} \cap \mathcal{X}_{c', t} \neq \emptyset$.

In order to model the arm group correlations, we maintain an undirected graph $\mathcal{G}_{t} = (V, E, W_{t})$ at each time step $t$, where each arm group from $\mathcal{C}$ is mapped to a corresponding node in node set $V$. Then, $E = \{e(c_{i}, c_{j})\}_{c_{i}, c_{j}\in \mathcal{C}}$ is the set of edges, and $W_{t}$ represents the set of edge weights. Note that by definition, $\mathcal{G}_{t}$ will stay as a fully-connected graph, and the estimated arm group correlations are modeled by the edge weights connecting pairs of nodes.
For a node $v \in V$, we denote the augmented $k$-hop neighborhood $\widetilde{\mathcal{N}}_{k}(v) = \mathcal{N}_{k}(v) \cup \{v\}$ as the union node set of its $k$-hop neighborhood $\mathcal{N}_{k}(v)$ and node $v$ itself. For the arm group graph $\mathcal{G}_{t}$, we denote $\matr{A}_{t} \in \mathbb{R}^{N_{c} \times N_{c}}$ as the adjacency matrix (with added self-loops) of the given arm group graph and $\matr{D}_{t} \in \mathbb{R}^{N_{c} \times N_{c}}$ as its degree matrix. For the notation consistency, we will apply a true arm group graph $\mathcal{G}^{*}$ instead of $\matr{W}^{*}$ in \textbf{Eq.} \ref{reward_func_defi} to represent the true arm group correlation.

\section{Proposed \name~Framework} \label{sec_proposed_framework}
In this section, we start with an overview to introduce our proposed \name~framework.
Then, we will show our estimation method of arm group graph before mentioning the related group-aware arm embedding. Afterwards, the two components of our proposed framework, namely the aggregation module and the reward-estimation module, will be presented.

\vspace{-0.2cm}
\subsection{Overview of \name~Framework}
In \textbf{Algorithm \ref{algo_1}}, we present the pseudo-code of proposed \name~framework.
%
\setlength{\textfloatsep}{10pt}  
\begin{algorithm}[t]
\caption{\name}
\label{algo_1}
\textbf{Input:} Number of rounds $T$, exploration parameter $\gamma$, regularization parameter $\lambda$, network width $m$, network depth $L$, neighborhood size $k$. \\
\textbf{Output:} Arm recommendation $\vect{x}_{t}$ for each time step $t$.\\
\textbf{Initialization:} Initialize the arm group graph as a connected graph $\mathcal{G}_{1} = (V, E, W_{1})$. Initialize gradient matrix $\matr{Z_{0}}=\lambda\matr{I}$. Initialize parameter $\vect{\Theta_{0}}$ for the model $f(\mathcal{G}, X;\vect{\Theta_{0}})$.\\
\For{$t = 1, 2, ..., T$}{
    Receive a set of arm contexts $\mathcal{X}_{t} = \{ \vect{x}^{(i)}_{c, t} \}_{c \in \mathcal{C}_{t}, i\in [n_{c, t}]}$.\\
    Embed the arm set $\mathcal{X}_{t}$ into $\widetilde{\mathcal{X}}_{t}$ w.r.t. \textbf{Eq.}\ref{new_embed_eq}.\\
    \For{each embedded arm $\widetilde{\matr{X}}^{(i)}_{c, t} \in \widetilde{\mathcal{X}}_{t}$}{
        Obtain the point estimate $\widehat{r}^{(i)}_{c, t} = f(\mathcal{G}_{t}, \widetilde{\matr{X}}^{(i)}_{c, t}; \vect{\Theta_{t-1}})$.\\
        Obtain network gradient $\vect{g}^{(i)}_{c, t} \xleftarrow{} \nabla_{\Theta}f(\mathcal{G}_{t}, \widetilde{\matr{X}}^{(i)}_{c, t}; \vect{\Theta_{t-1}})$. \\
        Calculate confidence bound as $\widehat{i}^{(i)}_{c, t} = \sqrt{\vect{g}_{c, t}^{(i)\intercal}\matr{Z}_{t-1}\vect{g}^{(i)}_{c, t} / m}$. \\
    }
    Recommend $\widetilde{\matr{X}}_{t} = \operatorname*{argmax}_{\widetilde{\matr{X}}^{(i)}_{c, t} \in \widetilde{\mathcal{X}}_{t}} (\widehat{r}^{(i)}_{c, t} + \gamma\cdot\widehat{i}^{(i)}_{c, t})$ with the received reward represented as $r_{t}$. \\
    Calculate arm group distances w.r.t. \textbf{Eq.}\ref{arm_similarity_eq}, and update the arm group graph $\mathcal{G}_{t}$ to $\mathcal{G}_{t+1}$. \\
    Update the model parameter $\vect{\Theta_{t-1}}$ to $\vect{\Theta_{t}}$ according to \textbf{Algorithm \ref{algo_2}}.\\
    Retrieve the $\widetilde{\matr{X}}_{t}$'s gradient vector $\vect{g}_{t}$, and update gradient matrix $\matr{Z}_{t} = \matr{Z}_{t-1}+\vect{g}_{t}\cdot\vect{g}_{t}^{\intercal}$. \\
}
\end{algorithm}
At each time step $t \in [T]$, \name~would receive a set of input arm contexts $\mathcal{X}_{t} = \{ \vect{x}^{(i)}_{c, t} \}_{c \in \mathcal{C}_{t}, i\in [n_{c, t}]}$ (line 5). Then, we embed the arm set $\mathcal{X}_{t}$ to $\widetilde{\mathcal{X}}_{t}$ based on \textbf{Eq.}\ref{new_embed_eq} from Subsection \ref{subsec_group_Aware_Arm_Embedding} (line 6).  
For each embedded arm $\widetilde{\matr{X}}\in\widetilde{\mathcal{X}}_{t}$, its estimated reward $\widehat{r}$ and confidence bound $\widehat{i}$ would be calculated (line 8-10) with the model $f(\cdot)$ in Subsection \ref{subsec_reward_est_module}. After recommending the best arm $\widetilde{\matr{X}}_{t}$ (line 12) and receiving its true reward $r_{t}$, we update the current arm group graph $\mathcal{G}_{t}$ based on Subsection \ref{subsec_arm_graph_est} (line 13).
Then, the model parameters $\vect{\Theta_{t-1}}$ will be trained based on \textbf{Algorithm \ref{algo_2}} (line 14), and we incrementally update the gradient matrix to $\matr{Z}_{t} = \matr{Z}_{t-1}+\vect{g}_{t}\cdot\vect{g}_{t}^{\intercal}$ with the gradient vector $\vect{g}_{t}$ of model $f(\cdot)$ given the selected arm $\widetilde{\matr{X}}_{t}$ (line 15).  

\begin{algorithm}[t]
\caption{Model Training}
\label{algo_2}
\textbf{Input:} Initial parameter $\vect{\Theta_{0}}$, step size $\eta$, training steps $J$, network width $m$. Updated arm group graph $\mathcal{G}_{t+1}$. Selected embedded contexts $\{\widetilde{\matr{X}}_{\tau}\}_{\tau=1}^{t}$.\\       
\textbf{Output:} Updated model parameter $\vect{\Theta_{t}}$.\\
$\matr{\Theta}_{t}^{0} \xleftarrow{} \matr{\Theta}_{0}$.\\
Let $\mathcal{L}(\matr{\Theta}) = \frac{1}{2}\sum_{\tau=1}^{t} \abs{f(\mathcal{G}_{t+1}, \widetilde{\matr{X}}_{\tau}; 
\matr{\Theta}) - r_{\tau}}^{2}$ \\
\For{$j = 1, 2, \dots, J$}{ 
    $\matr{\Theta}_{t}^{j} = \matr{\Theta}_{t}^{j-1} - \eta\cdot\nabla_{\matr{\Theta}} \mathcal{L}(\matr{\Theta}_{t}^{j-1})$ \\
} 
Return new parameter $\matr{\Theta}_{t}^{J}$. \\
\end{algorithm}
The steps from \textbf{Algorithm \ref{algo_2}} demonstrate our training process for \name~ parameters.
With the updated arm group graph $\mathcal{G}_{t+1}$ and the past embedded arm contexts $\{\widetilde{\matr{X}}_{\tau}\}_{\tau=1}^{t}$ until current time step $t$, we define the loss function as the straightforward quadratic loss function (line 4).
Finally, we run gradient descent (GD) for $J$ steps to derive the new model parameters $\Theta_{t}$ (lines 5-7) based on the initial parameters $\matr{\Theta}_{0}$ (initialized in Subsection \ref{subsec_model_init}).
Next, we proceed to introduce the detail of framework components.

\vspace{-0.2cm}
\subsection{Arm Group Graph Estimation}     \label{subsec_arm_graph_est}
Recall that at time step $t$, we model the similar arms into an arm group graph $\mathcal{G}_{t}=(V, E, W_{t})$ where the nodes $V$ are corresponding to the arm groups from $\mathcal{C}$ and edges weights $W_{t}$ formulate the correlations among arm groups. 
Given two nodes $\forall c, c' \in \mathcal{C}$,  to measure the similarity between them, inspired by the kernel mean embedding in the multi-task learning settings \cite{kernel_mean_2011,kmtl-ucb_2017}, we define edge weight between $c$ and $c'$ as: 
\begin{displaymath}
\begin{split}
    w^{*}(c, c') = \exp(-\norm{\mathbb{E}_{\vect{x}\sim \mathcal{D}_{c}}\big[\phi_{k_{\mathcal{G}}}(\vect{x})\big] - \mathbb{E}_{\vect{x}'\sim \mathcal{D}_{c'}}\big[\phi_{k_{\mathcal{G}}}(\vect{x}')\big]}^{2} / \sigma_{s})
\end{split}
\end{displaymath}
where $\phi_{k_{\mathcal{G}}}(\cdot)$ is the induced feature mapping of a given kernel $k_{\mathcal{G}}$, e.g., a radial basis function (RBF) kernel.
Unfortunately, $\forall c \in \mathcal{C}$, $\mathcal{D}_{c}$ is unknown. Therefore, we update the edge weight based on the empirical estimation of arm group correlations. 
Here, let $\mathcal{X}_{c}^{t} = \{\vect{x}^{(i)}_{c, \tau}\}_{\tau\in[t], i\in [n_{c, \tau}]}$ represent the set of all arm contexts from group $c\in\mathcal{C}$ up to time step $t$. We define the arm similarity measurement between arms $c, c' \in \mathcal{C}$ through a Gaussian-like kernel as
\begin{equation}
\begin{split}
    w_{t}(c, c') = \exp(-\norm{\Psi_{t}(\mathcal{D}_{c}) - \Psi_{t}(\mathcal{D}_{c'})}^{2} / \sigma_{s})
\end{split}
\label{arm_similarity_eq}
\end{equation}
where $\Psi_{t}(\mathcal{D}_{c})= \frac{1}{\abs{\mathcal{X}_{c}^{t}}} \sum_{x \in \mathcal{X}_{c}^{t}} k_{\mathcal{G}}(\cdot, x)$ denotes the kernel mean estimation of $\mathcal{D}_{c}$ with a given kernel $k_{\mathcal{G}}$; and $\sigma_{s}$ refers to the bandwidth. Then, at time step $t$ and $\forall c, c' \in \mathcal{C}$, we update the corresponding weight of edge $e(c, c')$ in the weight set $W_{t}$ with $w_{t}(c, c')$.

\subsection{Group-Aware Arm Embedding} \label{subsec_group_Aware_Arm_Embedding}
To conduct the aggregation operations of GNN, we reconstruct a matrix for each arm context vector.
Recall that for an arm group $c\in \mathcal{C}$, if $c \in \mathcal{C}_{t}$, we receive the contexts $x^{(i)}_{c, t} \in \mathbb{R}^{d_{x}}, i\in [n_{c, t}]$ at time step $t$. Then, the reconstructed matrix for $\vect{x}^{(i)}_{c, t}$ is defined as
\begin{equation}
\widetilde{\vect{X}}^{(i)}_{c, t} = 
\left(\begin{array}{cccc}
(\vect{x}^{(i)}_{c, t})^{\intercal} & \matr{0} & \cdots & \matr{0}\\
\matr{0} & (\vect{x}^{(i)}_{c, t})^{\intercal} & \cdots & \matr{0}\\
\vdots  &       & \ddots   & \vdots \\
\matr{0} & \matr{0} & \cdots & (\vect{x}^{(i)}_{c, t})^{\intercal}\\
\end{array}\right) \in \mathbb{R}^{N_{c}\times d_{\widetilde{x}}}
\label{new_embed_eq}
\end{equation}
where $d_{\widetilde{x}} = d_{x} \cdot N_{c}$ is the column dimension of $\widetilde{\vect{X}}^{(i)}_{c, t}$.
Here, for the $c'$-th row in matrix $\widetilde{\vect{X}}^{(i)}_{c, t}$, the $((c'-1)\cdot d_{x} + 1)$-th to the $(c'\cdot d_{x})$-th entries are the transposed original arm context $(\vect{x}^{(i)}_{c, t})^{\intercal}$, while the other entries are zeros. 
Receiving a set of arm contexts $\mathcal{X}_{t}$, we derive the corresponding embedded arm set as $\widetilde{\mathcal{X}}_{t} = \{ \widetilde{\vect{X}}^{(i)}_{c, t} \}_{c \in \mathcal{C}_{t}, i\in [n_{c, t}]}$. 



\subsubsection{Aggregation of arm group representations}

To leverage the estimated arm group graph for downstream reward estimations, we propose to aggregate over the arm group neighborhood for a more comprehensive arm representation through the GNN-based module, named as group-aware arm representation. 
It has been proven that the local averaging operation on the graph neighborhood can be deemed as applying the low-pass filter on the corresponding node features \cite{GNN_Low_pass_filter_hoang2021revisiting, SGC_wu2019simplifying}, which would give locally smooth node features within the same neighborhood. 
Inspired by the SGC model \cite{SGC_wu2019simplifying}, we propose to aggregate over the $k$-hop arm group neighborhood $\widetilde{\mathcal{N}}_{k}(\cdot)$ for incorporating arm group correlations to obtain the aggregated group-aware embedding for an embedded arm $\widetilde{\vect{X}}^{(i)}_{c, t}$, denoted by
\begin{equation}
\begin{split}
    \matr{H}_{gnn} = \sqrt{\frac{1}{m}}\cdot\sigma(\matr{S}_{t}^{k}\cdot \widetilde{\vect{X}}^{(i)}_{c, t} \matr{\Theta}_{gnn}) \in \mathbb{R}^{N_{c} \times m}
\end{split}
\label{aggr_Eq}
\end{equation}
where 
$\matr{S}_{t} = \matr{D}_{t}^{-\frac{1}{2}}\matr{A}_{t}\matr{D}_{t}^{-\frac{1}{2}}
$
is the symmetrically normalized adjacency matrix, and we have
\begin{displaymath}
\matr{\Theta}_{gnn} = 
\left(\begin{array}{c}
\matr{\Theta}_{gnn}^{1} \in \mathbb{R}^{d_{x}\times m}\\
\vdots \\
\matr{\Theta}_{gnn}^{c'} \in \mathbb{R}^{d_{x}\times m}\\
\vdots \\
\matr{\Theta}_{gnn}^{N_{c}} \in \mathbb{R}^{d_{x}\times m}\\
\end{array}\right) \in \mathbb{R}^{d_{\widetilde{x}}\times m}.
\end{displaymath}
being the trainable weight matrix with width $m$. Here, $\sigma(\cdot)$ denotes the non-linear activation function, which is added after the aggregation operation to alleviate potential concerns when the contexts are not linearly separable \cite{GNN_Low_pass_filter_hoang2021revisiting}.
Note that the $c'$-th row of $(\widetilde{\vect{X}}^{(i)}_{c, t}\cdot\matr{\Theta}_{gnn})$, denoted by $[\widetilde{\vect{X}}^{(i)}_{c, t} \cdot\matr{\Theta}_{gnn}]_{c',:}$, is the hidden representation of arm $\vect{x}$ in terms of $c'$-th arm group in $\mathcal{C}$.    
Then, these hidden representations will then be aggregated over $\widetilde{\mathcal{N}}_{k}(c), c\in \mathcal{C}$ by multiplying with $\matr{S}_{t}^{k}$ to derive the aggregated arm representation for $\vect{x}$, i.e.,  $\matr{H}_{gnn}(x)$.

\subsubsection{Incorporating initial embedded contexts}
Moreover, solely aggregating information from neighbors through the GNN-based models can lead to "over-smoothing" problems \cite{JK_Net_xu2018representation,linearized-GNN_xu2021optimization}. 
Aggregating from the node neighborhood will end up with identical representations for all the nodes if they form an isolated complete sub-graph,  which may not correctly reflect the relationship among these nodes in real-world applications. 
Therefore, we propose to apply skip-connections to address this potential problem by combining the initial contexts with the aggregated hidden features. 
Similar ideas have been applied to boost the performance of neural models. For instance, JK-Net \cite{JK_Net_xu2018representation} and GraphSAGE \cite{GraphSAGE_hamilton2017inductive} concatenate hidden features from different levels of node neighborhoods; and ResNet \cite{resnet_he2016deep} adopts additive residual connections.

Putting these two parts together and setting $d' = m~+~d_{\widetilde{x}}$, we then have
$\matr{H}_{0} \in \mathbb{R}^{N_{c} \times d'}$ as the output group-aware arm representation for $\widetilde{\vect{X}}^{(i)}_{c, t}$, represented by
\begin{equation}
\begin{split}
    \matr{H}_{0} = f_{gnn}(\mathcal{G}_{t}, \widetilde{\vect{X}}^{(i)}_{c, t}; \matr{\Theta}_{gnn}) = [\sigma(\matr{S}_{t}^{k}\cdot\widetilde{\vect{X}}^{(i)}_{c, t} \matr{\Theta}_{gnn}); \widetilde{\vect{X}}^{(i)}_{c, t}]
\end{split}
\label{SGC_eq}
\end{equation}
where $[\cdot~;~\cdot]$ refers to the column-wise concatenation of matrices.


\vspace{-0.2cm}
\subsection{Reward Estimation Module}   \label{subsec_reward_est_module}
In this subsection, we estimate the rewards with a FC network of $L$ layers and width $m$, based on group-aware arm representation $\matr{H}_{0}$. 

\subsubsection{Reward and confidence bound estimation}
Here, let $\matr{\Theta}_{fc} = \{\matr{\Theta}_{l}\}_{l \in [L]}$ be the set of trainable weight matrices of a fully-connected network,
where the specifications are: $\matr{\Theta}_{1} \in \mathbb{R}^{d' \times m }$, $\matr{\Theta}_{L} \in \mathbb{R}^{m}$ and $\matr{\Theta}_{l} \in \mathbb{R}^{m \times m}, \forall l \in \{2,\dots, L-1\}$. 
Then, given the group-aware representation $\matr{H}_{0}$,  
we have the reward estimation module as follows
\begin{equation}
\begin{split}
    &\matr{H}_{l} = \sqrt{\frac{1}{m}}\cdot\sigma( \matr{H}_{l-1} \cdot \matr{\Theta}_{l}  ),~~~~~ l \in \{1,\dots, L-1\}, \\
    &\widehat{\vect{r}}_{all} = f_{fc}(\matr{H}_{0}; \matr{\Theta}_{fc}) = \sqrt{\frac{1}{m}}\cdot \matr{H}_{L-1} \cdot \matr{\Theta}_{L}
\end{split}
\label{FC_model}
\end{equation}
where $\widehat{\vect{r}}_{all} \in \mathbb{R}^{N_{c}}$ represents the point-estimation vector for the received contexts embedding $\matr{H}_{0}$ with respect to all the arms groups. 
Given that the arm $\widetilde{\vect{x}}^{(i)}_{c, t}$ belonging to $c$-th group, 
 we will then have the reward estimation $\widehat{\vect{r}}^{(i)}_{c, t} = [\widehat{\vect{r}}_{all}]_{c}\in\mathbb{R}$ for the embedded context matrix $\widetilde{\vect{X}}^{(i)}_{c, t}$,
which is the $c$-th element of $\widehat{\vect{r}}_{all}$.



Finally, combining the aggregation module with the reward estimation module, given arm group graph $\mathcal{G}_{t}$ at time step $t$, the reward estimation for the embedded arm $\widetilde{\vect{X}}^{(i)}_{c, t}$ (i.e., the reward estimation given its arm group) can be represented as 
\begin{displaymath}
\begin{split}
    \widehat{r}^{(i)}_{c, t} = f(\mathcal{G}_{t}, \widetilde{\vect{X}}^{(i)}_{c, t}&; \matr{\Theta})= \Bigg[\bigg(f_{fc}(\cdot; \matr{\Theta}_{fc})\circ f_{gnn}(\cdot; \matr{\Theta}_{gnn})\bigg)(\mathcal{G}_{t}, \widetilde{\vect{X}}^{(i)}_{c, t}) \Bigg]_{c}.
\end{split}
\end{displaymath}
Setting $p = (2N_{c}\cdot d)\cdot m + (L-1)\cdot m^{2} + m$, we have $\matr{\Theta} \in \mathbb{R}^{p}$ being the set of all the parameters from these two modules.

\subsubsection{Arm pulling mechanism}
We obtain confidence bounds for the point estimation with the network gradients as 
$\widehat{i} = \sqrt{\vect{g}^{\intercal}\cdot\matr{Z}_{t-1}\cdot\vect{g} / m}$
where
$\vect{g} = \nabla_{\matr{\Theta}}f(\mathcal{G}_{t}, \widetilde{\vect{X}}^{(i)}_{c, t}; \matr{\Theta}) \in \mathbb{R}^{p}$ is the gradient vector, and $\matr{Z}_{t-1} = \matr{I}+\sum_{\tau=1}^{t-1}\vect{g}_{\tau}\cdot\vect{g}_{\tau}^{\intercal}$ with $\vect{g}_{\tau}$ being the gradient vector of the embedded arm which is selected at step $\tau\in\{1,\dots, t-1\}$.
After obtaining the reward and confidence bound estimations for all embedded arm in set $\widetilde{\mathcal{X}}_{t}$, we choose the best arm as
$\widetilde{\matr{X}}_{t} = \operatorname*{argmax}_{\widetilde{\matr{X}}^{(i)}_{c, t} \in \widetilde{\mathcal{X}}_{t}} (\widehat{r}^{(i)}_{c, t} + \gamma\cdot\widehat{i}^{(i)}_{c, t})$
where $\gamma$ is the exploration parameter, and the theoretical upper confidence bound will be given in Section \ref{sec_theoretical_analysis}.
Note that based on our problem definition (Section \ref{sec_problem_def}), one arm may associate with multiple arm groups. Here, we will separately estimate rewards and confidence bounds of each arm group it belongs to, and consider them as different arms for selection.

\vspace{-0.2cm}
\subsection{Model Initialization}   \label{subsec_model_init}
For the aggregation module weight matrix $\matr{\Theta}_{gnn}$, each of its entries is sampled from the Gaussian distribution $N(0, 1)$.
Similarly, the parameters from the first $L-1$ reward estimation module layers ($[\matr{\Theta}_{1}, \dots, \matr{\Theta}_{L-1}]$) are also sampled from $N(0, 1)$.
For the final ($L$-th) layer, its weight matrix $\matr{\Theta}_{L}$ is initialized by drawing the entry values from the Gaussian distribution $N(0, 1 / m)$.


\section{Theoretical Analysis} \label{sec_theoretical_analysis}
In this section, we provide the theoretical analysis for our proposed framework. For the sake of analysis, at each time step $t$, we assume each arm group $c\in \mathcal{C}$ would receive one arm $\vect{x}_{c, t}$,
which makes $\abs{\mathcal{X}_{1}^{t}} = \dots = \abs{\mathcal{X}_{N_{c}}^{t}} = t$.
We also apply the adjacency matrix $\matr{A}_{t}$ instead of $\matr{S}_{t}$ for aggregation, and set its elements
$[\matr{A}_{t}]_{ij} = \frac{1}{t\cdot N_{c}}\sum_{\tau=1}^{t}\phi_{k_{\mathcal{G}}}(x_{c_{i}, \tau})^{\intercal}\phi_{k_{\mathcal{G}}}(x_{c_{j}, \tau})$
for arm group similarity between group $c_{i}, c_{j}\in \mathcal{C}$. Here, $\phi_{k_{\mathcal{G}}}(\cdot)$ is the kernel mapping given an RBF kernel $k_{\mathcal{G}}$. With $\mathcal{G}^{*}$ being the unknown true arm group graph, its adjacency matrix elements are $[\matr{A}^{*}]_{ij} = \frac{1}{N_{c}}\mathbb{E}_{x_{i}\sim\mathcal{D}_{c_{i}}, x_{j}\sim\mathcal{D}_{c_{j}}} (\phi_{k_{\mathcal{G}}}(x_{i})^{\intercal}\phi_{k_{\mathcal{G}}}(x_{j}))$. Note that the norm of adjacency matrices $\norm{\matr{A}^{*}}_{2}, \norm{\matr{A}_{t}}_{2} \leq 1$ since $\inp{\phi_{k_{\mathcal{G}}}(x)}{\phi_{k_{\mathcal{G}}}(x')} \leq 1$ for any $x, x' \in \mathbb{R}^{d_{x}}$, which makes it feasible to aggregate over $k$-hop neighborhood without the explosion of eigenvalues.
Before presenting the main results, we first introduce the following background. 
\begin{lemma} [\cite{CNN_UCB-ban2021convolutional,Neural-UCB}]
For any $t \in [T]$, given arm $\vect{x} \in \mathbb{R}^{d_{x}}$ satisfying $\norm{\vect{x}}_{2}=1$ and its embedded context matrix $\widetilde{\matr{X}}$,  there exists $\matr{\Theta}_{t-1}^{*} \in \mathbb{R}^{p}$ at time step $t$, and a constant $S > 0$, such that
\begin{equation}
\begin{split}
    h(\mathcal{G}^{*}, \widetilde{\matr{X}}) = \inp{g(\mathcal{G}^{*}, \widetilde{\matr{X}};\matr{\Theta}_{t-1})}{\matr{\Theta}_{t-1}^{*} - \matr{\Theta}_{0}}
\end{split}
\label{eq_exp_reward}
\end{equation}
where $\norm{\matr{\Theta}_{t-1}^{*} - \matr{\Theta}_{0}}_{2} \leq S / \sqrt{m}$, $\forall t \in [T]$, and $\mathcal{G}^{*}$ stands for the unknown true underlying arm group graph. 
\label{lemma_expect_rewards}
\end{lemma}
Note that with sufficient network width $m$, we will have $\matr{\Theta}_{t-1}^{*} \approx \matr{\Theta}_{0}^{*},  \forall t \in [T]$, and we will include more details in the full version of the paper.
Following the analogous ideas from previous works \cite{Neural-UCB,neural_multifacet-ban2021multi}, this lemma formulates the expected reward as a linear function parameterized by the difference between randomly initialized network parameter $\matr{\Theta}_{0}$ and the parameter $\matr{\Theta}_{t-1}^{*}$, which lies in the confidence set with the high probability \cite{improved_linear_bandits_abbasi2011improved}. Then, regarding the activation function $\sigma(\cdot)$, we have the following assumption on its continuity and smoothness.
\begin{Assumption} [$\zeta$-Lipschitz continuity and Smoothness \cite{skip_kernel_du2019gradient,CNN_UCB-ban2021convolutional}]
For non-linear activation function $\sigma(\cdot)$, there exists a positive constant $\zeta > 0$, such that $\forall \vect{x}, \vect{x}' \in \mathbb{R}$, we have
\begin{displaymath}
\begin{split}
    &\abs{\sigma(x) - \sigma(x')} \leq \zeta \cdot \norm{x - x'}, \quad
    \abs{\sigma'(x) - \sigma'(x')} \leq \zeta \cdot \norm{x - x'}
\end{split}
\end{displaymath}
with $\sigma'(\cdot)$ being the derivative of activation function $\sigma(\cdot)$.
\label{assumption_act_func_Lipschitz}
\end{Assumption}
Note that \textbf{Assumption}~\ref{assumption_act_func_Lipschitz} is mild and applicable on many activation functions, such as Sigmoid.
Then, we proceed to bound the regret for a single time step $t$.

\subsection{Upper Confidence Bound}
Recall that at time step $t$, given an embedded arm matrix $\widetilde{\matr{X}}$, the output of our proposed framework is 
$\widehat{r} = f(\mathcal{G}_{t}, \widetilde{\matr{X}}; \matr{\Theta}_{t-1})$
with $\mathcal{G}_{t}$, $\matr{\Theta}_{t-1}$ as the estimated arm group graph and trained parameters respectively. The true function $h(\mathcal{G}^{*}, \widetilde{\matr{X}})$ is given in Lemma \ref{lemma_expect_rewards}. Supposing there exists the true arm group graph $\mathcal{G}^{*}$,
the confidence bound for a single round $t$ will be
\begin{equation}
\begin{split}
    & \textsf{CB}_{t}(\widetilde{\matr{X}}) = \abs{ f(\mathcal{G}_{t}, \widetilde{\matr{X}}; \matr{\Theta}_{t-1}) - h(\mathcal{G}^{*}, \widetilde{\matr{X}}) } \\
    &\leq \underbrace{\abs{ f(\mathcal{G}_{t}, \widetilde{\matr{X}}; \matr{\Theta}_{t-1}) - h(\mathcal{G}_{t}, \widetilde{\matr{X}}) }}_{R_{1}} + \underbrace{\abs{ h(\mathcal{G}_{t}, \widetilde{\matr{X}}) - h(\mathcal{G}^{*}, \widetilde{\matr{X}}) }}_{R_{2}}
\end{split}
\label{eq_regret_split}
\end{equation}
where $R_{1}$ denotes the error induced by network parameter estimations, and $R_{2}$ refers to the error from arm group graph estimations. We will then proceed to bound them separately.

\subsubsection{Bounding network parameter error $R_{1}$}
For simplicity, the $\mathcal{G}_{t}$ notation is omitted for this subsection.
To bridge the network parameters after GD with those at random initialization, we define the gradient-based regression estimator $\widehat{\matr{\Theta}}_{t} = \matr{Z}_{t}^{-1} \vect{b}_{t}$ where $\matr{Z}_{t} = \lambda\matr{I} + \frac{1}{m}\sum_{\tau=1}^{t} g(\widetilde{\matr{X}}_{\tau}; \matr{\Theta}_{\tau}) \cdot g(\widetilde{\matr{X}}_{\tau}; \matr{\Theta}_{\tau})^{\intercal}, 
\vect{b}_{t}  = \sum_{\tau=1}^{t} r_{\tau} \cdot g(\widetilde{\matr{X}}_{\tau}; \matr{\Theta}_{\tau}) / \sqrt{m}.$
Then, we derive the bound for $R_{1}$ with the following lemma.
\begin{lemma}
Assume there are constants $\beta_{F} > 0, ~~1 < \beta_{1}, \beta_{2}, \beta_{3}, \beta_{4} < 2,~~ \beta_{L} = \max\{\beta_{1}, \beta_{2}, \beta_{3}, \beta_{4}\}$, and
\begin{displaymath}
\begin{split}
    & \beta_{h} = \max\{\zeta \beta_{1},~~~ \zeta \beta_{2} + \zeta^{2} \beta_{1} \beta_{2},~~~ \zeta \beta_{L} + 1, 
    ~~~ (\zeta \beta_{4})^{L-2}(\zeta \beta_{2} + \zeta^{2} \beta_{1} \beta_{2})\}.
\end{split}
\end{displaymath}
With a constant $\delta\in (0, 1)$, and $L$ as the layer number for the FC network, let the network width  
$m \geq \text{Poly}\big( t, L, \frac{1}{\beta_{F}}, \frac{1}{\lambda}, (\zeta \beta_{L})^{L}, \log(\frac{1}{\delta}) \big)$, and learning rate $\eta \leq \mathcal{O}\big((t\cdot L \beta_{h}^{2}(2\zeta \beta_{L})^{2L})^{-1}\big)$.
Denoting the terms
\begin{displaymath}
\begin{split}
    & \Upsilon = \frac{2\sqrt{2}t}{\beta_{F}}(\beta_{h}+ \Lambda)  (\beta_{L}+ 1)^{L}  \zeta^{L}, 
    \quad\Lambda = \frac{\zeta \Upsilon \beta_{h}}{m} \cdot \frac{(2\zeta \beta_{L})^{L}-1}{2\zeta \beta_{L}-1} \\
    &\widetilde{I}_{1} = \sqrt{t\cdot (L\cdot \beta_{3}^{2}\cdot (\beta_{L}\zeta)^{2L} + m)} + \Lambda\cdot \sqrt{t\cdot(9L + m^{-1})}, \\
    & \widetilde{I}_{2} = \lambda \sqrt{L+1} \cdot \Upsilon / \sqrt{m},
\end{split}
\end{displaymath}
at time step $~t$, given the received contexts and rewards, with probability at least $1-\delta$ and the embedded context $\widetilde{\matr{X}}$, we have
\begin{displaymath}
\begin{split}
    & \abs{h(\widetilde{\matr{X}}) - f(\widetilde{\matr{X}}; \matr{\Theta}_{t-1})}
    \leq B_{1} \norm{g(\widetilde{\matr{X}}; \matr{\Theta}_{t-1}) / \sqrt{m}}_{\matr{Z}_{t-1}^{-1}} + B_{2} + B_{3}
\end{split}
\end{displaymath}
with the terms 
\begin{displaymath}
\begin{split}
    & B_{1} = \sqrt{\log(\frac{\det(\matr{Z}_{t-1})}{\det(\matr{\lambda I})}) - 2\log(\delta)} + \lambda^{\frac{1}{2}}S, \\
    & B_{2} = \big( 
        \frac{\widetilde{I}_{1} \cdot \sqrt{t} B_{3} +  m\cdot \widetilde{I}_{2}}{m\lambda} + \sqrt{\frac{t}{ m\lambda}} 
        \big) \\
        &\qquad 
        \cdot \big( 
        \Lambda\cdot\sqrt{9L + m^{-1}} + m^{-1}\beta_{h} \cdot \sqrt{L\cdot \beta_{3}^{2}\cdot (\beta_{L}\zeta)^{2L} + m} 
        \big) \\
    & B_{3} = m^{-0.5} \big(
        \beta_{3} (\Lambda+\beta_{h}) + L\cdot\Upsilon\cdot (\Lambda+\beta_{h})(\Lambda / \beta_{h} + 1) 
      \big).
\end{split}
\end{displaymath}

\label{lemma_CB_one_step_same_graph}
\end{lemma}

\textbf{Proof.} 
Given the embedded context $\widetilde{\matr{X}}$, and following the statement in Lemma \ref{lemma_expect_rewards}, we have 
\begin{displaymath}
\begin{split}
    & \abs{h(\widetilde{\matr{X}}) - f(\widetilde{\matr{X}}; \matr{\Theta}_{t-1})} \\
    & \quad\leq \abs{\inp{g(\widetilde{\matr{X}}; \matr{\Theta}_{t-1}) / \sqrt{m}}{\sqrt{m}(\matr{\Theta}_{t-1}^{*} - \matr{\Theta}_{0})} - \inp{g(\widetilde{\matr{X}}; \matr{\Theta}_{t-1}) / \sqrt{m}}{\widehat{\matr{\Theta}}_{t-1}}} \\
    &\quad\qquad + \abs{\inp{g(\widetilde{\matr{X}}; \matr{\Theta}_{t-1}) / \sqrt{m}}{\widehat{\matr{\Theta}}_{t-1}} - 
    f(\widetilde{\matr{X}}; \matr{\Theta}_{t-1})} = R_{3} + R_{4}.
\end{split}
\end{displaymath}
With Theorem 2 from \cite{improved_linear_bandits_abbasi2011improved}, we have $R_{3} \leq B_{1} \norm{g(\widetilde{\matr{X}}; \matr{\Theta}_{t-1}) / \sqrt{m}}_{\matr{Z}_{t-1}^{-1}}$.
Then, for $R_{4}$, we have $\abs{f(\widetilde{\matr{X}}; \matr{\Theta}_{t-1}) - \inp{g(\widetilde{\matr{X}}; \matr{\Theta}_{t-1} / \sqrt{m})}{\widehat{\matr{\Theta}}_{t-1}}} $
\begin{displaymath}
\begin{split}
   \leq  R_{5} + R_{6} & = \abs{f(\widetilde{\matr{X}}; \matr{\Theta}_{t-1}) - \inp{g(\widetilde{\matr{X}}; \matr{\Theta}_{t-1})}{\matr{\Theta}_{t-1} - \matr{\Theta}_{0}}} \\
    &\qquad + \abs{\inp{g(\widetilde{\matr{X}}; \matr{\Theta}_{t-1})}{\matr{\Theta}_{t-1} - \matr{\Theta}_{0} - \widehat{\matr{\Theta}}_{t-1} / \sqrt{m}}}
\end{split}
\end{displaymath}
where $R_{5}$ can be bounded by $B_{3}$ with Lemma \ref{lemma_output_minus_inner_product}. Then, with conclusions from Lemma \ref{lemma_linking_regre_est_with_net_param} and Lemma \ref{lemma_after_GD_gradient_for_network_norm}, we have
\begin{displaymath}
\begin{split}
     R_{6} &\leq \norm{\matr{\Theta}_{t-1} - \matr{\Theta}_{0} - \widehat{\matr{\Theta}}_{t-1} / \sqrt{m}}_{2}  \cdot \norm{g(\widetilde{\matr{X}}; \matr{\Theta}_{t-1})}_{2} \\
    & \leq B_{2} = \big( (\widetilde{I}_{1} \cdot \sqrt{t} B_{3} +  m\cdot \widetilde{I}_{2}) / (m\lambda) + \sqrt{t / (m\lambda)} \big) \\
    &\quad \cdot \big( \Lambda\cdot\sqrt{9L + m^{-1}} + m^{-1}\beta_{h} \cdot \sqrt{L\cdot \beta_{3}^{2}\cdot (\beta_{L}\zeta)^{2L} + m} \big),
\end{split}
\end{displaymath}
which completes the proof.      $\blacksquare$

\subsubsection{Bounding graph estimation error $R_{2}$} \label{subsection_R_2}
Regarding the regret term $R_{2}$ and for the aggregation module, we have
\begin{displaymath}
\begin{split}
    \matr{H}_{gnn} = \sqrt{\frac{1}{m}}\cdot\sigma(\matr{A}_{t}^{k}\cdot \widetilde{\matr{X}}\matr{\Theta}_{gnn}) \in \mathbb{R}^{N_{c} \times m}
\end{split}
\end{displaymath}
as the output where $\matr{\Theta}_{gnn}$ refers to the trainable weight matrix. 
Then, we use the following lemma to bound $R_{2}$.
\begin{lemma}
At this time step $t+1$, given any two arm groups $c_{i}, c_{j} \in \mathcal{C}$ and their sampled arm contexts $\mathcal{X}_{c_{i}}^{t} = \{\vect{x}_{c_{i}, \tau}\}_{\tau=1}^{t}$, $\mathcal{X}_{c_{j}}^{t} = \{\vect{x}_{c_{j}, \tau}\}_{\tau=1}^{t}$,
with the notation from Lemma \ref{lemma_CB_one_step_same_graph} and the probability at least $1-\delta$, we have
\begin{displaymath}
\begin{split}
    \norm{\matr{A}^{*} - \matr{A}_{t}}_{\max} \leq \frac{1}{N_{c}}\cdot\sqrt{\frac{1}{2t}\log(\frac{N_{c}^{2} - N_{c}}{\delta})}
\end{split}
\end{displaymath}
where $\norm{\cdot}_{\max}$ refers to the greatest entry of a matrix. Then, we will have $R_{2} \leq B_{4} \sqrt{1 / t}$ with
\begin{displaymath}
\begin{split}
    B_{4} = \frac{ S\sqrt{L} k }{\sqrt{m}} (\beta_{h} + \Lambda) (\zeta \beta_{L} + \frac{\Upsilon\zeta}{m})^{O(L)} \sqrt{\frac{1}{2}\log(\frac{N_{c}^{2} - N_{c}}{\delta})},
\end{split}
\end{displaymath}
and $N_{c} = \abs{\mathcal{C}}$ is the number of arm groups.
\label{lemma_adjacency_matrix_concentration}
\end{lemma}
\textbf{Proof.}
Recall that for $c_{i}, c_{j} \in \mathcal{C}$, the element of matrix
$[\matr{A}^{*}]_{ij} = \\ \frac{1}{N_{c}}\mathbb{E}_{x_{i}\sim\mathcal{D}_{c_{i}}, x_{j}\sim\mathcal{D}_{c_{j}}} (\phi_{k_{\mathcal{G}}}(x_{i})^{\intercal}\phi_{k_{\mathcal{G}}}(x_{j})), \forall i, j \in [N_{c}]$, and $[\matr{A}_{t}]_{ij} = \frac{1}{t\cdot N_{c}}\sum_{\tau=1}^{t}\phi_{k_{\mathcal{G}}}(x_{c_{i}, \tau})^{\intercal}\phi_{k_{\mathcal{G}}}(x_{c_{j}, \tau})$. Here, suppose a distribution $\mathcal{D}_{ij}$ where $\mathbb{E}[\mathcal{D}_{ij}] =  \frac{1}{N_{c}}\mathbb{E}_{x_{i}\sim\mathcal{D}_{c_{i}}, x_{j}\sim\mathcal{D}_{c_{j}}} (\phi_{k_{\mathcal{G}}}(x_{i})^{\intercal}\phi_{k_{\mathcal{G}}}(x_{j}))$. 
Given $N_{c}$ arm groups, we have $N_{c}(N_{c} - 1) / 2$ different group pairs. For group pair $c_{i}, c_{j}\in \mathcal{C}$, each $\phi_{k_{\mathcal{G}}}(x_{c_{i}, \tau})^{\intercal}\phi_{k_{\mathcal{G}}}(x_{c_{j}, \tau}), \tau\in [t]$ is a sample drawn from $\mathcal{D}_{ij}$, and the element distance $\abs{[\matr{A}_{t}]_{ij} - [\matr{A}^{*}]_{ij}}$ can be regarded as the difference between the mean value of samples and the expectation. Applying the Hoeffding's inequality and the union bound would complete the proof.
As $\norm{\cdot}_{2} \leq n \norm{\cdot}_{\max}$ for an $n\times n$ square matrix, we have the bound for matrix differences.

Then, consider the power of adjacency matrix $\matr{A}^{k}$ (for graph $\mathcal{G}$) as input and fix $\widetilde{\matr{X}}$. Analogous to the idea that the activation function with the Lipschitz continuity and smoothness property will lead to Lipschitz neural networks \cite{conv_theory-allen2019convergence}, applying Assumption \ref{assumption_act_func_Lipschitz} and with Lemma \ref{lemma_after_GD_weight_matrices_bounds}, Lemma \ref{lemma_after_GD_model_results_variables}, we simply have the gradient $g(\mathcal{G}, \widetilde{\matr{X}};\matr{\Theta}_{t-1})$ being Lipschitz continuous w.r.t. the input graph as
\begin{displaymath}
\begin{split}
    R_{2} & \leq 
    \norm{g(\mathcal{G}^{*}, \widetilde{\matr{X}};\matr{\Theta}_{t-1}) - g(\mathcal{G}_{t}, \widetilde{\matr{X}};\matr{\Theta}_{t-1})}_{2} \cdot \norm{\matr{\Theta}_{t-1}^{*} - \matr{\Theta}_{0}}_{2}  \\
    & \leq \frac{S\sqrt{L} }{\sqrt{m}} (\beta_{h} + \Lambda) (\zeta \beta_{L} + \frac{\Upsilon\zeta}{\sqrt{m}})^{O(L)}\cdot \abs{\norm{(\matr{A}_{t})^{k}}_{2}  - \norm{\matr{(A^{*}})^{k}}_{2}} \\
    & \underset{(i)}{\leq} \frac{S\sqrt{L} k}{\sqrt{m}} (\beta_{h} + \Lambda) (\zeta \beta_{L} + \frac{\Upsilon\zeta}{\sqrt{m}})^{O(L)}\cdot \norm{\matr{A}_{t} - \matr{A^{*}}}_{2}
\end{split}
\end{displaymath}
where $(i)$ is because $\matr{A}_{t}, \matr{A}^{*}$ are symmetric and bounded polynomial functions are Lipschitz continuous.
Combining the two parts will lead to the conclusion.
$\blacksquare$

\subsubsection{Combining $R_{2}$ with $R_{1}$}
At time step $t$, with the notation and conclusions from Lemma \ref{lemma_CB_one_step_same_graph} and Lemma \ref{lemma_adjacency_matrix_concentration}, re-scaling the constant $\delta$, we have the confidence bound given embedded arm $\widetilde{\matr{X}}$ as
\begin{equation}
\begin{split}
    \textsf{CB}_{t}(\widetilde{\matr{X}}) \leq  B_{1} \norm{g(\mathcal{G}_{t}, \widetilde{\matr{X}}; \matr{\Theta}_{t-1}) / \sqrt{m}}_{\matr{Z}_{t-1}^{-1}} + B_{2} + B_{3} + B_{4} \sqrt{\frac{1}{t}}.
\end{split}
\label{eq_CB_one_step}
\end{equation}

\vspace{-0.2cm}
\subsection{Regret Bound}

With the UCB shown in Eq. \ref{eq_CB_one_step}, we provide the following regret upper bound $R(T)$, for a total of $T$ time steps.

\begin{theorem}
Given the received contexts and rewards, with the notation from Lemma \ref{lemma_CB_one_step_same_graph}, Lemma \ref{lemma_adjacency_matrix_concentration}, and probability at least $1-\delta$, if $m, \eta$ satisfy conditions in Lemma \ref{lemma_CB_one_step_same_graph}, we will have the regret
\begin{displaymath}
\begin{split}
    & R(T) \leq 2\cdot (2B_{4}\sqrt{T} + 2 - B_{4}) + 2\sqrt{2\widetilde{d} T \log(1+T/\lambda) + 2T} \\
    &\qquad \cdot \big(\sqrt{\lambda}S + \sqrt{1-2\log(\delta / 2) + (\widetilde{d} \log(1+T/\lambda))}\big) 
\end{split}
\end{displaymath}
where the effective dimension $\widetilde{d} = \frac{\log\det(\matr{I} + \matr{G}(0) / \lambda)}{\log(1+T/\lambda)}$
with \\
$\matr{G}(0) = \matr{G}_{0}\matr{G}_{0}^{\intercal}$  
and $\matr{G}_{0} = \big( g(\widetilde{\matr{X}}_{1}; \matr{\Theta}_{0})^{\intercal}, \dots, g(\widetilde{\matr{X}}_{t}; \matr{\Theta}_{0})^{\intercal} \big)$. 
\label{theorem_regret_bound}
\end{theorem}

\textbf{Proof.}
By definition, we have the regret $R_{t}$ for time step $t$ as
\begin{displaymath}
\begin{split}
    R_{t} &= 
    h(\mathcal{G}^{*}, \widetilde{\matr{X}}_{t}^{*}) - h(\mathcal{G}^{*}, \widetilde{\matr{X}}_{t}) \\
    & \leq \textsf{CB}_{t}(\widetilde{\matr{X}}_{t}^{*}) +  f(\mathcal{G}_{t}, \widetilde{\matr{X}}_{t}^{*}; \matr{\Theta}_{t-1})
    - h(\mathcal{G}^{*}, \widetilde{\matr{X}}_{t}) \\
    & \leq \textsf{CB}_{t}(\widetilde{\matr{X}}_{t}) +  f(\mathcal{G}_{t}, \widetilde{\matr{X}}_{t}; \matr{\Theta}_{t-1})
    - h(\mathcal{G}^{*}, \widetilde{\matr{X}}_{t})
    \leq 2\cdot \textsf{CB}_{t}(\widetilde{\matr{X}}_{t})
\end{split}
\end{displaymath}
where the second inequality is due to our arm pulling mechanism.
Then, based on Lemma \ref{lemma_adjacency_matrix_concentration}, Lemma \ref{lemma_CB_one_step_same_graph}, and Eq. \ref{eq_CB_one_step}, we have $R(T) =$
\begin{displaymath}
\begin{split}
    \sum_{t=1}^{T} & R_{t}   \leq  2\sum_{t=1}^{T} \bigg(B_{1} \norm{g(\mathcal{G}_{t},  \widetilde{\matr{X}}; \matr{\Theta}_{t-1}) / \sqrt{m}}_{\matr{Z}_{t-1}^{-1}} + B_{2} + B_{3} + B_{4} \sqrt{\frac{1}{t}} \bigg)\\
    & \leq 2\cdot (2B_{4}\sqrt{T} + 2 - B_{4}) + 2\sum_{t=1}^{T} (B_{1} \norm{g(\mathcal{G}_{t},  \widetilde{\matr{X}}; \matr{\Theta}_{t-1}) / \sqrt{m}}_{\matr{Z}_{t-1}^{-1}})
\end{split}
\end{displaymath}
with the choice of $m$ for bounding the summation of $B_{2}, B_{3}$, and the bound of $\sum_{i=1}^{T}[t^{-i / 2}]$ in \cite{bound_sum_sqrt_T_chlebus2009approximate}. Then, with Lemma 11 from \cite{improved_linear_bandits_abbasi2011improved}, 
\begin{displaymath}
\begin{split}
    & \sum_{t=1}^{T} (B_{1} \norm{g(\mathcal{G}_{t},  \widetilde{\matr{X}}; \matr{\Theta}_{t-1}) / \sqrt{m}}_{\matr{Z}_{t-1}^{-1}}) \\
    & \leq B_{1} \sqrt{T\sum_{t=1}^{T}  \norm{g(\mathcal{G}_{t},  \widetilde{\matr{X}}; \matr{\Theta}_{t-1}) / \sqrt{m}}_{\matr{Z}_{t-1}^{-1}}^{2}}
     \leq \sqrt{T} B_{1} \sqrt{2\log(\frac{\det(\matr{Z}_{T})}{\det(\lambda \matr{I})})} \\
    & \underset{(i)}{\leq} \sqrt{2\widetilde{d} T \log(1+T/\lambda) + 2T} 
     \big(\sqrt{\lambda}S + \sqrt{1-2\log(\delta / 2) + (\widetilde{d} \log(1+T/\lambda))}\big)
\end{split}
\end{displaymath}
where $(i)$ is based on Lemma 6.3 in \cite{CNN_UCB-ban2021convolutional} and Lemma 5.4 in \cite{Neural-UCB}. 
$\blacksquare$

Here, the effective dimension $\widetilde{d}$ measures the vanishing speed of $\matr{G}(0)$'s eigenvalues, and it is analogous to that of existing works on neural contextual bandits algorithms \cite{CNN_UCB-ban2021convolutional,Neural-UCB,neural_multifacet-ban2021multi}. As $\widetilde{d}$ is smaller than the dimension of the gradient matrix $\matr{G}(0)$, it is applied to prevent the dimension explosion. Our result matches the state-of-the-art regret complexity \cite{Neural-UCB,neural_thompson-zhang2020neural,CNN_UCB-ban2021convolutional} under the worst-case scenario.

\vspace{-0.2cm}
\subsection{Model Convergence after GD}
For model convergence, we first give an assumption of the gradient matrix after $j$ iterations of GD. First, we define $\matr{G}^{(j)}(\matr{\Theta}_{L-1}) =
\big( g(\widetilde{\matr{X}}_{1}; \matr{\Theta}_{L-1}^{(j)}), \dots, g(\widetilde{\matr{X}}_{T}; \matr{\Theta}_{L-1}^{(j)}) \big)^{\intercal}
\big( g(\widetilde{\matr{X}}_{1}; \matr{\Theta}_{L-1}^{(j)}), \dots, g(\widetilde{\matr{X}}_{T}; \matr{\Theta}_{L-1}^{(j)}) \big)$ \\
where $g(\widetilde{\matr{X}}; \matr{\Theta}_{L-1})$ is the gradient vector w.r.t. $\matr{\Theta}_{L-1}$.
\begin{Assumption}
With width $m \geq \mbox{Poly}(T, L, \frac{1}{\beta_{F}}, \frac{1}{\lambda}, (\zeta \beta_{L})^{L}, \log(\frac{1}{\delta}))$ and for $j \in [J]$, we have the minimal eigenvalue of $\matr{G}^{(j)}$ as
    \begin{displaymath}
    \begin{split}
        \lambda_{\min}(\matr{G}^{(j)}(\matr{\Theta}_{L-1})) \geq \lambda_{0} / 2
    \end{split}
    \end{displaymath}
where $\lambda_{0}$ is the minimal eigenvalue of the neural tangent kernel (NTK) \cite{NTK_jacot2018neural} matrix induced by \name.
\label{assumption_gradient_matrix_eigenvalue}
\end{Assumption}
Note that Assumption \ref{assumption_gradient_matrix_eigenvalue} is mild and has been proved for various neural architectures in \cite{skip_kernel_du2019gradient}. 
The NTK for \name~ can be derived following a comparable approach as in \cite{GNTK_du2019graph,NTK_jacot2018neural}.
Then, we apply the following lemma and theorem to prove the convergence of \name.
The proof of Lemma \ref{lemma_after_GD_next_iteration_output} is given in the appendix.


\begin{lemma}
After $T$ time steps, assume the network are trained with the $J$-iterations GD on the past contexts and rewards. Then, with  $\beta_{F} > 0$ and $\beta_{F}\cdot \eta < 1$, for any $j\in [J]$:
\begin{displaymath}
\begin{split}
    \norm{\vect{F}^{(j)}_{T} - \vect{F}^{(j+1)}_{T}}_{2}^{2} \leq \frac{1}{4}\eta \beta_{F} \cdot \norm{\vect{F}^{(j)}_{T} - \vect{Y}_{T}}_{2}^{2}
\end{split}
\end{displaymath}
with network width $m$ defined in Lemma \ref{lemma_CB_one_step_same_graph}.
\label{lemma_after_GD_next_iteration_output}
\end{lemma}
The Lemma \ref{lemma_after_GD_next_iteration_output} shows that we are able to bound the difference in network outputs after one step of GD. Then, we proceed to prove the convergence with the theorem below.

\begin{theorem}
After $T$ time steps, assume the model with width $m$ defined in Lemma \ref{lemma_CB_one_step_same_graph} is trained with the $J$-iterations GD on the contexts $\{\widetilde{\matr{X}}_{\tau}\}_{\tau=1}^{T}$ and rewards $\{r_{\tau}\}_{\tau=1}^{T}$. With probability at least $1 - \delta$, a constant $\beta_{F}$ such that $\beta_{F}\cdot \eta < 1$, set the network width $m \geq \mbox{Poly}(T, L, \frac{1}{\beta_{F}}, \frac{1}{\lambda}, (\zeta \beta_{L})^{L}, \log(\frac{1}{\delta}))$ and the learning rate $\eta \leq \mathcal{O}(T^{-1}L^{-1}\beta_{h}^{-2}(2\zeta \beta_{L})^{-2L})$. Then, for any $j\in [J]$, we have
\begin{displaymath}
\begin{split}
    \norm{\vect{F}^{(j)}_{T}- \vect{Y}_{T}}_{2}^{2} \leq (1 - \beta_{F}\cdot \eta)^{j} \cdot \norm{\vect{F}^{(0)}_{T}- \vect{Y}_{T}}_{2}^{2}
\end{split}
\end{displaymath}
where the vector $\vect{F}^{(j)} = [f(\mathcal{G}_{T}, \widetilde{\matr{X}}_{\tau};\matr{\Theta}^{(j)})]_{\tau=1}^{T}$, and $\vect{Y}_{T} = [r_{\tau}]_{\tau=1}^{T}$.
\label{theorem_after_GD_param_outputs}
\end{theorem}

\textbf{Proof.}
Following an approach analogous to \cite{skip_kernel_du2019gradient}, we apply and induction based method for the proof. The hypothesis is that $\norm{\vect{F}^{(j)}_{T}- \vect{Y}_{T}}_{2}^{2} \leq (1 - \beta_{F}\cdot \eta)^{j} \cdot \norm{\vect{F}^{(0)}_{T}- \vect{Y}_{T}}_{2}^{2}, j\in [J]$. With a similar procedure in Condition A.1 of \cite{skip_kernel_du2019gradient}, we have
\begin{displaymath}
\begin{split}
    & \norm{\vect{F}^{(j+1)}_{T}- \vect{Y}_{T}}_{2}^{2} \leq 
    \norm{\vect{F}^{(j)}_{T}- \vect{Y}_{T}}_{2}^{2} - 2\eta\norm{\vect{F}^{(j)}_{T}- \vect{Y}_{T}}_{\matr{G}^{(j)}}^{2} \\
    &\qquad - 2(\vect{Y}_{T} - \vect{F}^{(j)}_{T})^{\intercal} \matr{V}^{(j)} + 
    \norm{\vect{F}^{(j+1)}_{T}- \vect{F}^{(j)}_{T}}_{2}^{2}
\end{split}
\end{displaymath}
with $\matr{V}^{(j)} = (\matr{V}^{(j)}(\widetilde{\matr{X}}_{1}), \dots, \matr{V}^{(j)}(\widetilde{\matr{X}}_{T}))^{\intercal}$. For $\matr{\Theta}' \in \{\matr{\Theta}_{gnn}, \dots, \matr{\Theta}_{L-1}\}$,
\begin{displaymath}
\begin{split}
    & \abs{\matr{V}^{(j)}(\widetilde{\matr{X}})} = 
    \eta \max_{0\leq s\leq \eta} \bigg[ \sum_{\matr{\Theta}'} 
    \norm{\nabla\mathcal{L}({\matr{\Theta}'}^{(j)})}_{F} 
    \norm{\nabla f({\matr{\Theta}'}^{(j)}) - \nabla f({\matr{\Theta}'}^{(j)}, s)}_{F}
    \bigg]
\end{split}
\end{displaymath}
where $\nabla f({\matr{\Theta}'}^{(j)}, s) =
\nabla f \big( {\matr{\Theta}'}^{(j)} - s\cdot \nabla\mathcal{L}({\matr{\Theta}'}^{(j)}) \big)$. The notation $\mathcal{G}, \widetilde{\matr{X}}$ is omitted for simplicity.
Then, based on the conclusions from Lemma \ref{lemma_after_GD_I_2_term}, Lemma \ref{lemma_after_GD_next_iteration_output} and Assumption \ref{assumption_gradient_matrix_eigenvalue}, we can have
\begin{displaymath}
\begin{split}
    & \norm{\vect{F}^{(j+1)}_{T}- \vect{Y}_{T}}_{2}^{2} \leq 
    (1-\eta\lambda_{0})\norm{\vect{F}^{(j)}_{T} - \vect{Y}_{T}}_{2}^{2} - 2(\vect{Y}_{T} - \vect{F}^{(j)}_{T})^{\intercal} \matr{V}^{(j)} \\
    & \qquad+ \norm{\vect{F}^{(j+1)}_{T}- \vect{F}^{(j)}_{T}}_{2}^{2}
    \leq (1-\frac{\eta\lambda_{0}}{2})\norm{\vect{F}^{(j)}_{T} - \vect{Y}_{T}}_{2}^{2}
\end{split}
\end{displaymath}
by setting $\beta_{F} = \lambda_{0} / 2$.         $\blacksquare$

This theorem shows that with sufficiently large $m$ and proper $\eta$, the GD will converge to the global minimum at a linear rate, which is essential for proving the regret bound.

\vspace{-0.1cm}
\section{Experiments} \label{sec_experiments}
In this section, we demonstrate the effectiveness of our proposed framework by comparing its performances with state-of-the-art baselines through experiments on four real data sets. As linear algorithms have been outperformed in previous works \cite{Neural-UCB,neural_thompson-zhang2020neural,kmtl-ucb_2017}, we will not include these linear methods in the experiments below.  
Our six baseline algorithms are: 
\begin{itemize}
    \item KMTL-UCB \cite{kmtl-ucb_2017} estimates the "task similarities" with received contextual information. The estimations are based on a variant of kernel ridge regression.
    \item Kernel-Ind is Kernel-UCB \cite{kernel_ucb-2013} under the \textit{"disjoint setting"} \cite{linucb-Ind_li2010contextual} where it learns individual estimators for each arm group. 
    \item Kernel-Pool represents Kernel-UCB under the \textit{"pooling setting"} where it applies a single estimator for all arm groups.
    \item Neural-TS stands for Neural Thompson Sampling \cite{neural_thompson-zhang2020neural} 
    with group-aware embedding, which enables it to leverage the group information. It applies a neural network for exploitation and Thompson sampling strategy for exploration.
    \item Neural-Pool is for Neural-UCB \cite{Neural-UCB} with a single neural network to evaluate the reward, and calculate the upper confidence bounds with the network gradients. 
    \item Neural-Ind represents Neural-UCB with group-aware embedding for utilizing the group information. 
\end{itemize}
%

Note that COFIBA \cite{co_filter_bandits_2016} is naturally Kernel-Ind (with linear kernel) given the arm group information and one single user to serve, so we do not include it in our benchmarks.
To find the best exploration parameter, we perform grid searches over the range $\{10^{-1}, 10^{-2}, 10^{-3}\}$ for all algorithms. Similarly, the learning rate for neural algorithms are chosen from $\{10^{-2}, 10^{-3}, 10^{-4}\}$. For Neural-UCB, Neural-TS and our reward estimation module, we apply a two-layer FC network with $m=500$. RBF kernels are applied for KMTL-UCB and Kernel-UCB as well as our graph estimation module. Kernel-Pool and Neural-Pool will not fit into the multi-class classification setting, as we only receive one arm (context) at each time step without the arm group information.

\begin{figure}[t]
  \centering
  \includegraphics[width=\linewidth]{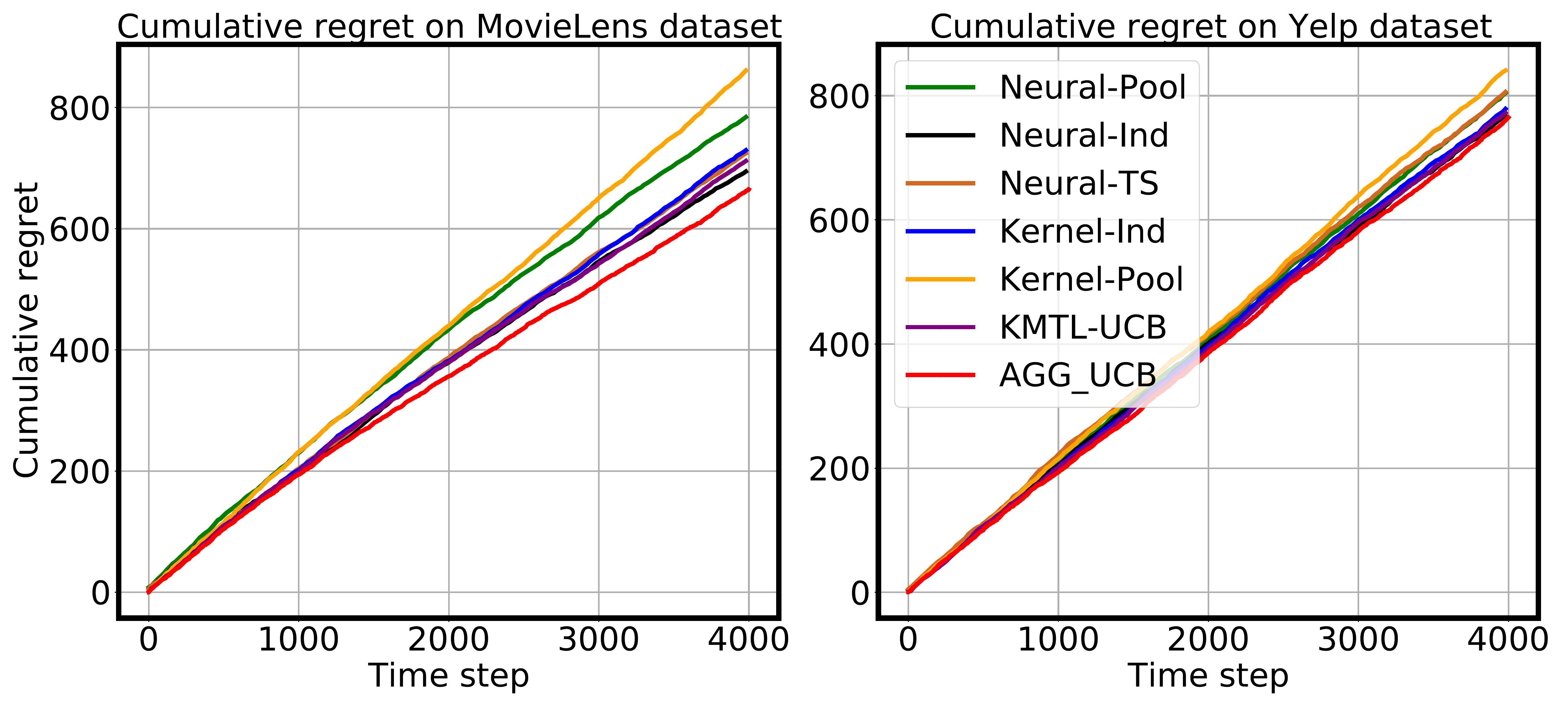}
  \vspace{-0.7cm}
  \caption{Cumulative regrets for recommendation data sets.}
  \label{figure_recommendation_result}
  \vspace{-0.1cm}
\end{figure}

\vspace{-0.2cm}
\subsection{Real Data Sets}
Here, we compare our proposed model with baseline algorithms on four real data sets with different specifications. 

\textit{MovieLens and Yelp data sets.}
The first real data set is the "MovieLens 20M rating data set" 
(\url{grouplens.org/datasets/movielens/20m/})
. 
To obtain the user features, we first choose 100 movies and 4000 users with \textbf{most reviews} to form the user-movie matrix where the entries are user ratings, and the user features $\vect{v}_{u}\in\mathbb{R}^{d}$ are obtained through singular value decomposition (SVD) where the dimension $d=20$. Then, since the  genome-scores of user-specified tags are provided for each movie, we select 20 tags with the highest variance to construct the movie features $\vect{v}_{i} \in \mathbb{R}^{d}$ with their scores on these tags. Then, these movies are allocated into 19 groups based on their genres ($\abs{\mathcal{C}}=19$).
Receiving a user $u_{t}$ at each time step $t$, we follow the idea of Generalized Matrix Factorization (GMF) \cite{NCF_he2017neural,PURE_zhou2021pure,XRS_Yao_2021} to encode user information into the contexts as $\widetilde{\vect{x}}^{(i)}_{c, t} = [\vect{v}_{u_{t}} \odot \vect{v}_{i}] \in \mathbb{R}^{d}, c\in \mathcal{C}_{t}, i\in[n_{c, t}]$, and let $\abs{\mathcal{X}_{t}} = 20$. Finally, we concatenate a constant 0.01 to each $\widetilde{\vect{x}}^{(i)}_{c, t}$ to obtain $\vect{x}^{(i)}_{c, t} \in \mathbb{R}^{d_{x}}$, which makes $d_{x}=21$, before normalizing $\vect{x}^{(i)}_{c, t}$. Rewards $r_{c, t}^{(i)}$ are user ratings normalized into range [0, 1]. 

Then, for the Yelp data set (\url{https://www.yelp.com/dataset}), we choose 4000 users with \textbf{most reviews} and restaurants from 20 different categories as arms $(\abs{\mathcal{C}}=20)$. Both user features and arm features are obtained through SVD with the dimension $d=20$. Analogous to the MovieLens data set, we follow the GMF based approach and the fore-mentioned constant concatenation to get the arm context $\vect{x}^{(i)}_{c, t}$ ($d_{x}=21, \abs{\mathcal{X}_{t}}=20$) to encode the user information, and the rewards are the normalized user ratings.

\textit{MNIST data set with augmented classes (MNIST-Aug)}.
MNIST is a well-known classification data set with 10 original classes where each sample is labeled as a digit from 0 to 9. Here, we further divide the samples from each class into 5 sub-divisions through $K$-means clustering, which gives us a total of 50 augmented sub-classes (i.e., arm groups) for the whole data set. Given a sample $\vect{x}_{t}$, the reward would be $r_{t}=1$ if the learner accurately predicts its sub-class; or the learner will receive the partial reward $r_{t}=0.5$ when it chooses the wrong sub-class, but this sub-class and the correct one belong to the same digit (original class). Otherwise, the reward $r_{t}=0$.

\textit{XRMB data set}.
XRMB data set \cite{XRMB_wang2015unsupervised} is a multi-view classification data set with 40 different labels. Here, we only apply samples from the first 38 classes as there are insufficient samples for the last two classes. The arm contexts $\vect{x}_{t}$ are the first-view features of the samples. Then, learner will receive a reward of $r_{t}=1$ when they predict the right label, and $r_{t}=0$ otherwise.

\vspace{-0.2cm}
\subsection{Experimental Results}


\begin{figure}[t]
  \centering
  \includegraphics[width=\linewidth]{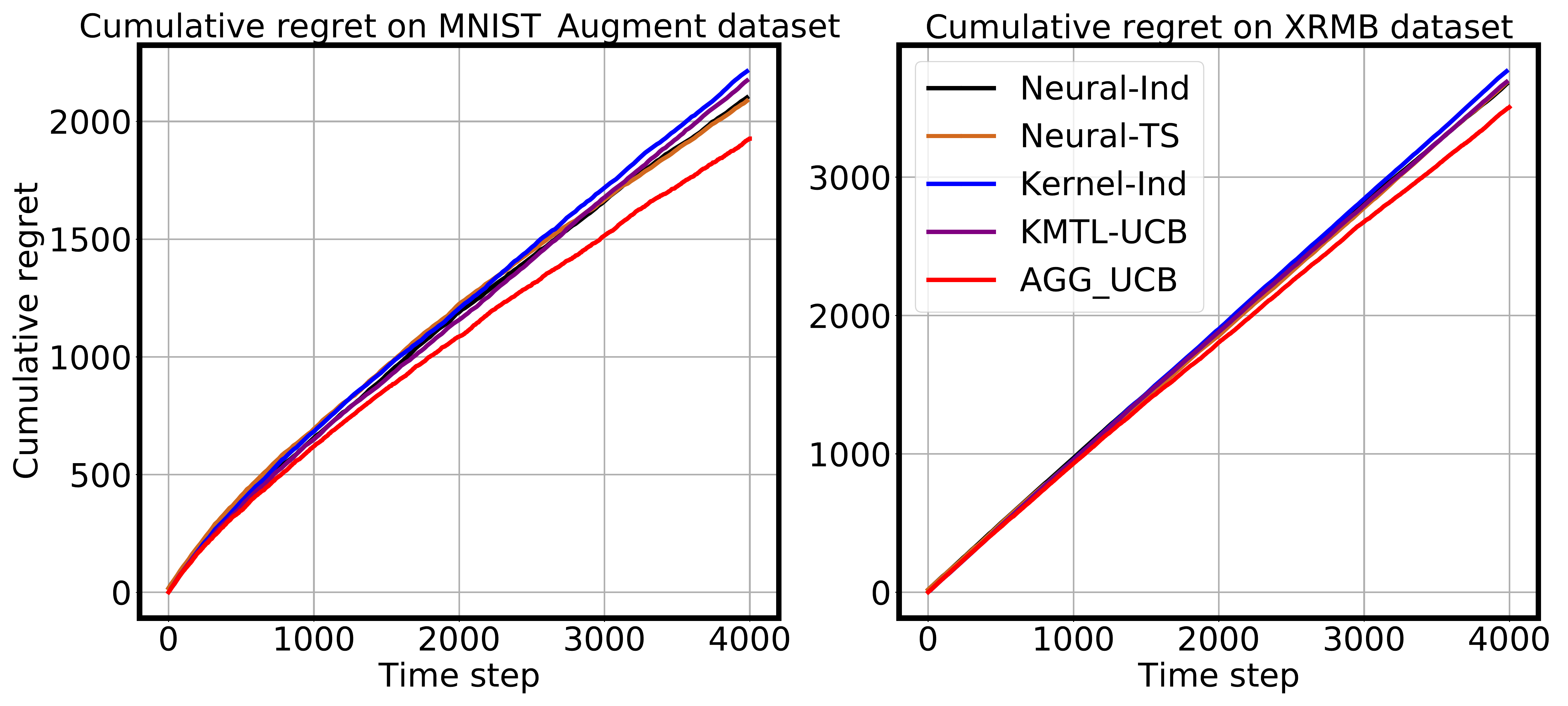}
  \vspace{-0.7cm}
  \caption{Cumulative regrets for classification data sets.}
  \label{figure_classification_result}
  \vspace{-0.2cm}
\end{figure}



Figure \ref{figure_recommendation_result} shows the cumulative regret results on the two real recommendation data sets where our proposed \name~ outperforms all strong baselines. In particular, we can find that algorithms with group-aware arm embedding tend to perform better than those without the arm group information (Kernel-Pool, Neural-Pool). This confirms the necessity of exploiting arm group information. Nevertheless, these baselines fed with group-aware are outperformed by \name, which implies the advantages of of our new graph-based model.
Meantime, it can be observed that neural algorithms (\name, Neural-Ind, Neural-TS) generally perform better compared with other baselines due to the representation power of neural networks. Note that since the user features and arm features of the Yelp data set are directly extracted with SVD, the reward estimation on the Yelp data set is comparably easy compared with others data sets. Therefore, the performances of benchmarks do not differ dramatically with \name. In opposite, MovieLens data set with true arm features tends to be a more challenging task where a more complex mapping from arms to their rewards can be involved. This can be reason for \name's superiority over the competitors.

Then, Figure \ref{figure_classification_result} shows the cumulative regret results on the two classification data sets where our \name~ achieves the best performance compared with other baselines. In particular, since sub-classes from each digit are highly correlated in the MNIST-Aug data set, our proposed \name~ tends to perform significantly better due to its ability of leveraging arm group correlations compared with other neural methods. Thus, these two aspects verify our claim that associating the neural models with arm group relationship modeling can lead to better performance.

\vspace{-0.2cm}
\subsection{Parameter Study}

\begin{figure}[t]
  \centering
  \includegraphics[width=\linewidth]{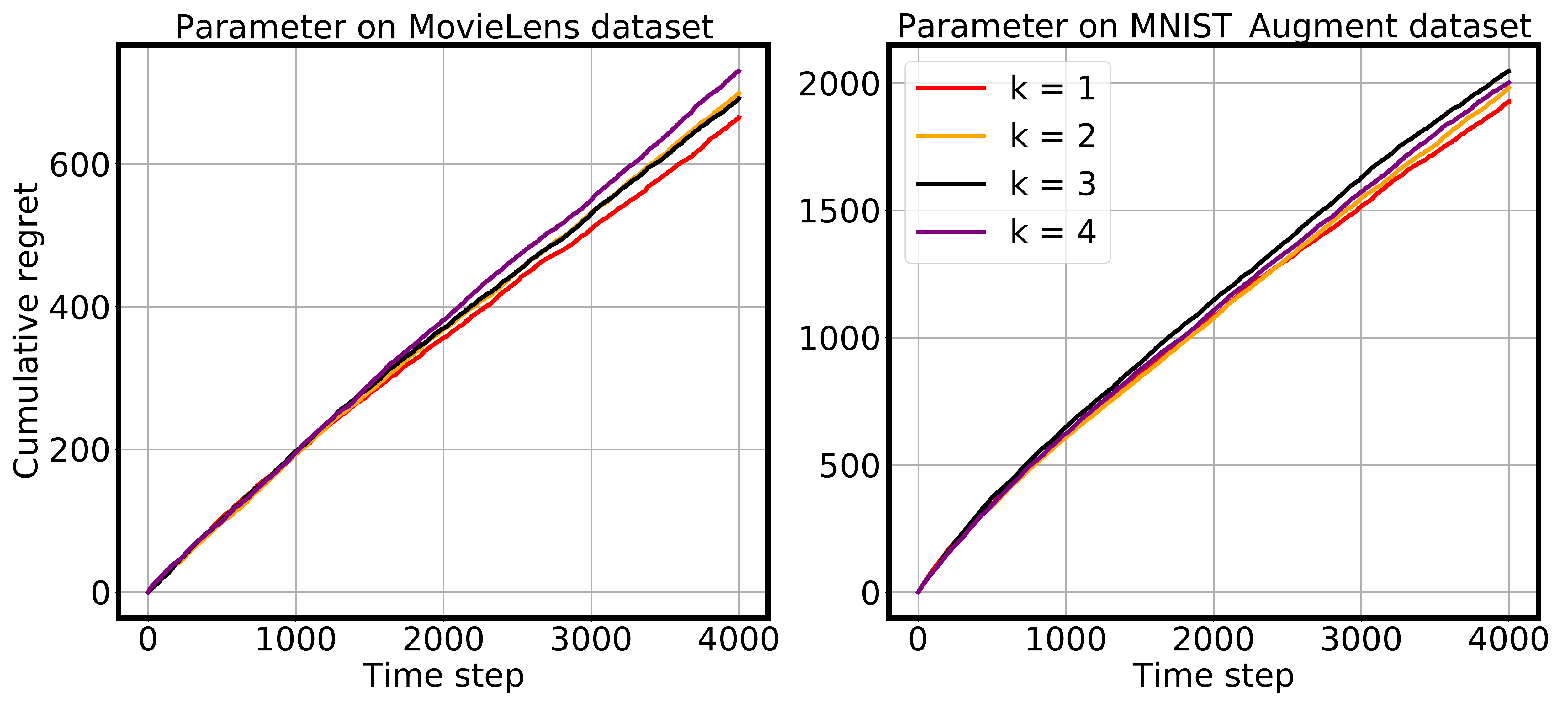}
  \vspace{-0.7cm}
  \caption{Cumulative regrets on MovieLens and MNIST-Aug data sets with different neighborhood parameter $k$.}
  \label{figure_parameter_study}
  \vspace{-0.2cm}
\end{figure}

In this section, we conduct our parameter study for the neighborhood parameter $k$ on the MovieLens data set and MNIST-Aug data set with augmented labels, and the results are presented in Figure \ref{figure_parameter_study}. For the MovieLens data set, we can observe that setting $k=1$ would give the best result. Although increasing $k$ can enable the aggregation module to propagate the hidden representations for multiple hops, it can potentially fail to focus on local arm group neighbors with high correlations, which is comparable to the aforementioned "over-smoothing" problem. In addition, since the arm group graph of MovieLens data set only has 19 nodes, $k=1$ would be enough. Meantime, setting $k=1$ also achieves the best performance on the MNIST data set. The reason can be that the $1$-hop neighborhood of each sub-class can already include all the other sub-classes from the same digit with heavy edge weights within the neighborhood for arm group collaboration. Therefore, unless setting $k$ to considerably large values, the \name~can maintain robust performances, which reduces the workload for hyperparameter tuning.

\vspace{-0.2cm}
\section{Conclusion} \label{sec_conclusion}
In this paper, motivated by real applications where the arm group information is available, we propose a new graph-based model to characterize the relationship among arm groups.
Base on this model, we propose a novel UCB-based algorithm named \name, which uses GNN to exploit the arm group relationship and share the information across similar arm groups. 
Compared with existing methods, \name~ provides a new way of collaborating multiple neural contextual bandit estimators for obtaining the rewards.
In addition to the theoretical analysis of \name, we empirically demonstrate its superiority on real data sets in comparison with state-of-the-art baselines.


\begin{acks}
This work is supported by National Science Foundation under Award No. IIS-1947203, IIS-2117902, IIS-2137468, and IIS-2002540. The views and conclusions are those of the authors and should not be interpreted as representing the official policies of the funding agencies or the government.
\end{acks}


%
\bibliographystyle{ACM-Reference-Format}
\bibliography{AGG_UCB}

\clearpage
\appendix

\section{Lemmas for Intermediate Variables and Weight Matrices} \label{sec_appendix_A}
Due to page limit, we will give the proof sketch for lemmas at the end of each corresponding appendix section. 
Recall that each input context $x^{(i)}_{c, t}, i\in [n_{c, t}]$ is embedded to $\widetilde{\matr{X}}^{(i)}_{c, t}$ (represented by $\widetilde{\matr{X}}$ for brevity).
Supposing $\widetilde{\matr{X}}$ belongs to the arm group $c$, denote $\vect{h}_{\matr{A}} = [\matr{A}_{t}^{k}\widetilde{\matr{X}}]_{c}$ as the corresponding row in matrix $\matr{A}_{t}^{k}\widetilde{\matr{X}}$ based on index of group $c$ in $\mathcal{C}$ (if group $c$ is the $c'$-th group in $\mathcal{C}$, then $\vect{h}_{\matr{A}}$ is the $c'$-th row in $\matr{A}_{t}^{k}\widetilde{\matr{X}}$). 
Similarly, we have $\vect{h}_{gnn} = [\matr{H}_{gnn}]_{c}$ and $\vect{h}_{l} = [\matr{H}_{l}]_{c}$ respectively. 
Given received contexts $\{\widetilde{\matr{X}}_{\tau}\}_{\tau=1}^{T}$ and rewards $\{r_{\tau}\}_{\tau=1}^{T}$, the gradient w.r.t. weight matrix $\matr{\Theta}_{l}, \forall l \in \{1, \dots, L-1\}$ will be
\begin{displaymath}
\begin{split}
    \frac{\partial~\mathcal{L}(\Theta)}{\partial~\matr{\Theta}_{l}} = m^{-\frac{L-l+1}{2}}\sum_{\tau=1}^{T} 
    \abs{f(\widetilde{\matr{X}}_{\tau};\matr{\Theta}) - r_{\tau}}^{2}  \bigg( \vect{h}_{l-1} \matr{\Theta}_{L}^{\intercal} 
    \big(\prod_{q=l+1}^{L-1} \matr{\Gamma}_{q} \matr{\Theta}_{q}^{\intercal} \big)
     \matr{\Gamma}_{l} 
    \bigg)
\end{split}
\end{displaymath}
where $\matr{\Gamma}_{q} = diag([\sigma'(\vect{h}_{q-1}\matr{\Theta}_{q})])$ is the diagonal matrix whose entries are the elements from $\sigma'(\matr{\vect{h}_{q-1}\Theta}_{q})$. The coefficient $\frac{1}{2}$ of the cost function is omitted for simplicity.
Then, for $\matr{\Theta}_{gnn}$, we have
\begin{displaymath}
\begin{split}
    \frac{\partial~\mathcal{L}(\Theta)}{\partial~\matr{\Theta}_{gnn}} = m^{-\frac{L+1}{2}}\sum_{\tau=1}^{T} 
    \abs{f(\widetilde{\matr{X}}_{\tau};\matr{\Theta}) - r_{\tau}}^{2}  \bigg(\vect{h}_{\matr{A}} \matr{\Theta}_{L}^{\intercal} 
    \big(\prod_{q=2}^{L-1} \matr{\Gamma}_{q} \matr{\Theta}_{q}^{\intercal} \big) \matr{\Gamma}_{1} \matr{\Theta}_{1}^{\intercal}
      \matr{Q} \matr{\Gamma}_{gnn} \bigg)
\end{split}
\end{displaymath}
where $\matr{\Gamma}_{gnn} = diag([\sigma'(\vect{h}_{\matr{A}} \matr{\Theta}_{gnn})])$.
$\matr{Q} = $
$\left(
\begin{array}{c} 
  \matr{I}\in \mathbb{R}^{m\times m} \\ 
  \hline 
  \matr{0} \in \mathbb{R}^{(d' - m)\times m}
\end{array} 
\right) \in \mathbb{R}^{d'\times m}$. 
Given the same $\mathcal{G}_{t}$, we provide lemmas to bound the $R_{1}$ term of \textbf{Eq.} \ref{eq_regret_split}.
For brevity, the subscript $\tau\in [T]$ and notation $\mathcal{G}_{t}$ are omitted below by default.

\begin{lemma}
Given the randomly initialized parameters $\matr{\Theta}^{(0)} = \{\matr{\Theta}_{gnn}^{(0)}, \matr{\Theta}_{1}^{(0)}, \matr{\Theta}_{2}^{(0)}, \dots, \matr{\Theta}_{L}^{(0)}\}$, with the probability at least $1 - O(TL)\cdot e^{-\Omega(m)}$ and constants $1 < \beta_{1}, \beta_{2}, \beta_{3}, \beta_{4} < 2$, we have
\begin{displaymath}
\begin{split}
    &\norm{\matr{\Theta}_{gnn}^{(0)}}_{2} \leq \beta_{1}\sqrt{m},~ 
    \norm{\matr{\Theta}_{1}^{(0)}}_{2} \leq \beta_{2}\sqrt{m},~
    \norm{\matr{\Theta}_{L}^{(0)}}_{2} \leq \beta_{3}, \\
    &\norm{\vect{h}_{gnn}^{(0)}}_{2} \leq \zeta\cdot \beta_{1}, \quad
    \norm{\vect{h}_{1}^{(0)}}_{2} \leq \zeta\cdot \beta_{2} + \zeta^{2}\cdot \beta_{1} \beta_{2}, \\
    &\abs{f(\widetilde{\matr{X}};\matr{\Theta}^{(0)})} \leq \zeta \cdot \beta_{3}\cdot(\zeta\cdot \beta_{4})^{L-2}(\zeta\cdot \beta_{2} + \zeta^{2}\cdot \beta_{1} \beta_{2}) / \sqrt{m}, \\
    &\norm{\matr{\Theta}_{l}^{(0)}}_{2} \leq \beta_{4}\sqrt{m},~~ \norm{\vect{h}_{l}^{(0)}}_{2} \leq (\zeta\cdot \beta_{4})^{l-1}(\zeta\cdot \beta_{2} + \zeta^{2}\cdot \beta_{1} \beta_{2}),~~  \\
    &\forall l \in \{2, \dots, L-1\}.
\end{split}
\end{displaymath}

\label{lemma_initial_param_outputs}
\end{lemma}


\textbf{Proof.}
Based on the properties of random Gaussian matrices \cite{random_matrix-vershynin2010introduction, skip_kernel_du2019gradient, CNN_UCB-ban2021convolutional}, with the probability of at least $1 - e^{-\frac{(\beta_{1} - \sqrt{d_{\widetilde{x}} / m} - 1)^{2}\cdot m}{2}} = 1 - e^{-\Omega(m)}$, we have
\begin{displaymath}
\begin{split}
    \norm{\matr{\Theta}_{gnn}^{(0)}}_{2} \leq \beta_{1}\sqrt{m}
\end{split}
\end{displaymath}
where $\beta_{1} \geq \sqrt{d_{\widetilde{x}} / m} + 1$ with $m > d_{\widetilde{x}}$.
Applying the analogous approach for the other randomly initialized matrices would give similar bounds.
Regarding the nature of $\matr{A}$, we can easily have $\norm{\vect{h}_{\matr{A}}}_{2} \leq 1$. Then,
\begin{displaymath}
\begin{split}
    \norm{\vect{h}_{gnn}^{(0)}}_{2} = m^{-\frac{1}{2}}\norm{\sigma(\vect{h}_{\matr{A}}\cdot\matr{\Theta}_{gnn})}_{2} \leq \zeta m^{-\frac{1}{2}}\cdot \norm{\vect{h}_{\matr{A}}}_{2} \norm{\matr{\Theta}_{gnn}}_{2} \leq \zeta\cdot \beta_{1}
\end{split}
\end{displaymath}
due to the assumed $\zeta$-Lipschitz continuity. Denoting the concatenated input for reward estimation module as $\vect{x}' = [\vect{h}_{gnn}^{(0)}; \widetilde{\matr{X}}]_{c} \in \mathbb{R}^{1\times (d_{\widetilde{x}} + m)}$, we can easily derive that $\norm{\vect{x}'}_{2} \leq \zeta\cdot \beta_{1} + 1$. Thus,
\begin{displaymath}
\begin{split}
    \norm{\vect{h}_{1}^{(0)}}_{2}& = m^{-\frac{1}{2}}\norm{\sigma(\vect{x}'\cdot\matr{\Theta}_{1})}_{2} 
    \leq \zeta m^{-\frac{1}{2}}\cdot \norm{\vect{x}'}_{2}\norm{\matr{\Theta}_{1}}_{2} \\ 
    & \leq \zeta\cdot \beta_{2} (\zeta\cdot \beta_{1} + 1) =  \zeta\cdot \beta_{2} + \zeta^{2}\cdot \beta_{1} \beta_{2}.
\end{split}
\end{displaymath}
Following the same procedure recursively for other intermediate outputs and applying the union bound would complete the proof.

\begin{lemma}
After $T$ time steps, run GD for $J$-iterations on the network with the received contexts and rewards. Suppose $\norm{\vect{h}_{l}^{(j)} - \vect{h}_{l}^{(0)}}_{2} \leq \Lambda^{(j)}, \forall j \in [J]$. With the probability of at least $1 - O(TL)\cdot e^{-\Omega(m)}$ and $\matr{\Theta}\in \{\matr{\Theta}_{gnn}, \matr{\Theta}_{1}, \dots, \matr{\Theta}_{L}\}$, we have
\begin{displaymath}
\begin{split}
    &\norm{\matr{\Theta}^{(j)} - \matr{\Theta}^{(0)}}_{F} \leq \Upsilon / \sqrt{m}
\end{split}
\end{displaymath}
where
$
    \Upsilon = \frac{2\sqrt{2}t}{\beta_{F}}(\beta_{h}+ \Lambda^{(j)})  (\beta_{L}+ 1)^{L}  \zeta^{L}.
$
\label{lemma_after_GD_weight_matrices_bounds}
\end{lemma}



\textbf{Proof.}
We prove this Lemma following an induction-based procedure \cite{skip_kernel_du2019gradient, CNN_UCB-ban2021convolutional}. The hypothesis is $\norm{\matr{\Theta}^{(0)} - \matr{\Theta}^{(j)}}_{F} \leq \Upsilon/\sqrt{m}, \forall \matr{\Theta}\in \{\matr{\Theta}_{gnn}, \matr{\Theta}_{1}, \dots, \matr{\Theta}_{L}\}$, and let $\beta_{L} = \max\{\beta_{1}, \beta_{2}, \beta_{3}, \beta_{4}\}$.
According to \textbf{Algorithm}~\ref{algo_2}, we have for the $j+1$-th iteration and $l \in [L]$, 
\begin{displaymath}
\begin{split}
    &\norm{\matr{\Theta}_{l}^{(j+1)} - \matr{\Theta}_{l}^{(j)}}_{F} 
    =m^{-\frac{L-l+1}{2}} \eta\cdot\norm{\sum_{\tau=1}^{T}
    \abs{f(\widetilde{\matr{X}}_{\tau};\matr{\Theta}^{(j)}) - r_{\tau}}^{2} \cdot 
    \vect{h}_{l-1}^{(j)} (\matr{\Theta}_{L}^{(j)})^{\intercal}  \\
    &\qquad\cdot\big(\prod_{q=l+1}^{L-1} \matr{\Gamma}_{q}^{(j)}  \cdot (\matr{\Theta}_{q}^{(j)})^{\intercal} \big)
    \cdot \matr{\Gamma}_{l}^{(j)}
    }_{F} \\
    &\leq m^{-\frac{L-l+1}{2}}\eta\sqrt{t}\cdot\norm{\vect{F}^{(j)}_{t}- \vect{Y}_{t}}_{2}
    \norm{\vect{h}_{l-1}^{(j)}\matr{\Theta}_{L}^{(j)}}_{F} 
    \prod_{q=l+1}^{L-1}\norm{\matr{\Theta}_{q}^{(j)}}_{2} 
    \prod_{q=l}^{L-1} \norm{\matr{\Gamma}_{q}^{(j)}}_{2} \\
    &\leq m^{-\frac{L-l+1}{2}}\eta\sqrt{t}\cdot\norm{\vect{F}^{(j)}_{t}- \vect{Y}_{t}}_{2}
    \norm{\vect{h}_{l-1}^{(j)}}_{2} \norm{\matr{\Theta}_{L}^{(j)}}_{2}  
    \prod_{q=l+1}^{L-1}\norm{\matr{\Theta}_{q}^{(j)}}_{2} 
    \prod_{q=l}^{L-1} \norm{\matr{\Gamma}_{q}^{(j)}}_{2}
\end{split}
\end{displaymath}
by Cauchy inequality.
For $\norm{\matr{\Theta}_{q}^{(j)}}_{2}$, we have
\begin{displaymath}
\begin{split}
    \prod_{q=l+1}^{L-1}\norm{\matr{\Theta}_{q}^{(j)}}_{2} &\leq \prod_{q=l+1}^{L-1}\bigg(\norm{\matr{\Theta}_{q}^{(0)}}_{2} + \norm{\matr{\Theta}_{q}^{(j)} - \matr{\Theta}_{q}^{(0)}}_{2}\bigg)\\
    &\qquad \leq (\beta_{L} \sqrt{m}+ \Upsilon/\sqrt{m})^{L-l-1};
\end{split}
\end{displaymath}
while for $\norm{\matr{\Gamma}_{q}^{(j)}}_{2}$, we have $\prod_{q=l}^{L-1} \norm{\matr{\Gamma}_{q}^{(j)}}_{2} \leq \zeta^{L-l}$. Combining all the results above and based on Lemma \ref{theorem_after_GD_param_outputs}, it means that for $l \in [L]$, 
\begin{displaymath}
\begin{split}
    &\norm{\matr{\Theta}_{l}^{(j+1)} - \matr{\Theta}_{l}^{(j)}}_{F} 
    \leq m^{-\frac{L-l+1}{2}}\eta\sqrt{t}\cdot(1 - \beta_{F}\cdot \eta)^{j / 2} \cdot \norm{\vect{F}^{(0)}_{t}- \vect{Y}_{t}}_{2} \\
    &\qquad\cdot \norm{\vect{h}_{l-1}^{(j)}}_{2}\cdot\norm{\matr{\Theta}_{L}^{(j)}}_{2}\cdot(\beta_{L} \sqrt{m}+ \Upsilon/\sqrt{m})^{L-l-1} \cdot \zeta^{L-l+1} \\
    &\leq m^{-\frac{1}{2}}(1 - \beta_{F} \eta)^{j / 2}\eta\sqrt{t} \norm{\vect{F}^{(0)}_{t}- \vect{Y}_{t}}_{2}
     ((\beta_{h}+ \Lambda^{(j)})  (\beta_{L}+ \Upsilon/m)^{L-l}  \zeta^{L-l}
\end{split}
\end{displaymath}   
where the last inequality is due to Lemma~\ref{lemma_after_GD_model_results_variables}. Then, since we have $\norm{\matr{\Theta}_{l}^{(j+1)} - \matr{\Theta}_{l}^{(j)}}_{F} \leq \norm{\matr{\Theta}_{l}^{(j+1)} - \matr{\Theta}_{l}^{(j)}}_{F} + \norm{\matr{\Theta}_{l}^{(j)} - \matr{\Theta}_{l}^{(0)}}_{F}$, it leads to
\begin{displaymath}
\begin{split}
    &\norm{\matr{\Theta}_{l}^{(j+1)} - \matr{\Theta}_{l}^{(0)}}_{F} 
    \leq \frac{2\sqrt{t}}{\beta_{F}\sqrt{m}}\norm{\vect{F}^{(0)}_{t}- \vect{Y}_{t}}_{2}
     (\beta_{h}+ \Lambda^{(j)})  (\beta_{L}+ \Upsilon/m)^{L-l}  \zeta^{L-l}.
\end{split}
\end{displaymath}   
For the last layer $\matr{\Theta}_{L}$, the conclusion can be verified through a similar procedure. 
Analogously, for $\matr{\Theta}_{gnn}$, we have
\begin{displaymath}
\begin{split}
    &\norm{\matr{\Theta}_{gnn}^{(j+1)} - \matr{\Theta}_{gnn}^{(j)}}_{F} \\
    & = m^{-\frac{L+1}{2}}\eta \norm{\sum_{\tau=1}^{T} 
    \abs{f(\widetilde{\matr{X}}_{\tau};\matr{\Theta}) - r_{\tau}}^{2} \cdot \bigg(\vect{h}_{\matr{A}}\cdot \matr{\Theta}_{L}^{\intercal} \cdot
    \big(\prod_{q=2}^{L-1} \matr{\Gamma}_{q} \cdot \matr{\Theta}_{q}^{\intercal}\big)
    \cdot \matr{\Gamma}_{1} \matr{\Theta}_{1}^{\intercal}
    \bigg)\\
    &\qquad\qquad \cdot \matr{Q} \cdot \matr{\Gamma}_{gnn}}_{F} \\
    &\leq m^{-\frac{L+1}{2}}\eta\sqrt{t}(1 - \beta_{F}\cdot \eta)^{j/2}  \norm{\vect{F}^{(0)}_{t}- \vect{Y}_{t}}_{2} \norm{\vect{h}_{\matr{A}}}_{2}\norm{\prod_{q=1}^{L} \matr{\Theta}_{q}}_{2}  \zeta^{L} \norm{\matr{Q}}_{2} \\
    &\leq \sqrt{t} m^{-\frac{1}{2}}(1 - \beta_{F}\cdot \eta)^{j/2}\eta\cdot \norm{\vect{F}^{(0)}_{t}- \vect{Y}_{t}}_{2} \cdot \zeta^{L} \cdot(\beta_{L}+ \Upsilon/m)^{L},
\end{split}
\end{displaymath}
which leads to 
\begin{displaymath}
\begin{split}
    &\norm{\matr{\Theta}_{gnn}^{(j+1)} - \matr{\Theta}_{gnn}^{(0)}}_{F} 
    \leq \frac{2}{\beta_{F}} \sqrt{t} m^{-\frac{1}{2}}\cdot \norm{\vect{F}^{(0)}_{t}- \vect{Y}_{t}}_{2} \cdot \zeta^{L} \cdot(\beta_{L}+ \Upsilon/m)^{L}.
\end{split}
\end{displaymath}
Since $\norm{\vect{F}^{(0)}_{t} - \vect{Y}_{t}}_{2}\leq \sqrt{2t}$ (Lemma \ref{theorem_after_GD_param_outputs}) and $\Upsilon/m \leq 1$ with sufficiently large $m$, combining all the results above would give the conclusion.

\begin{lemma}
After $T$ time steps, with the probability of at least $1 - O(TL)\cdot e^{-\Omega(m)}$ and running GD of $J$-iterations on the contexts and rewards, we have $\beta'_{h} = \max\{\zeta\cdot \beta_{1}, \zeta\cdot \beta_{2} + \zeta^{2}\cdot \beta_{1} \beta_{2}, (\zeta\cdot \beta_{4})^{L-2}(\zeta\cdot \beta_{2} + \zeta^{2}\cdot \beta_{1} \beta_{2})\}$ and $\beta_{h}=\max\{\zeta\cdot \beta_{L} + 1, \beta'_{h}\}$.
With $\vect{h} \in \{\vect{h}_{gnn}, \vect{h}_{1}, \dots, \vect{h}_{L-1}\}$, we have
\begin{displaymath}
\begin{split}
    &\norm{\vect{h}^{(j)} - \vect{h}^{(0)}}_{2} \leq \frac{\zeta \Upsilon}{m}\cdot \beta_{h} \cdot \frac{(2\zeta \beta_{L})^{L}-1}{2\zeta \beta_{L}-1} = \Lambda^{(j)},~
    \norm{\vect{h}^{(j)}}_{2} \leq \beta_{h} + \Lambda^{(j)}
\end{split}
\end{displaymath}

\label{lemma_after_GD_model_results_variables}
\end{lemma}


\textbf{Proof.}
Similar to the proof of \textbf{Lemma}~\ref{lemma_after_GD_weight_matrices_bounds}, we adopt an induction-based approach. For $l \in [L-1]$, we have 
\begin{displaymath}
\begin{split}
    &\norm{\vect{h}_{l}^{(j)} - \vect{h}_{l}^{(0)}}_{2} = \sqrt{\frac{1}{m}} \norm{\sigma( \vect{h}_{l-1}^{(j)}\cdot \matr{\Theta}_{l}^{(j)} ) - \sigma(\vect{h}_{l-1}^{(0)}\cdot \matr{\Theta}_{l}^{(0)})}_{2} \\
    & \leq \sqrt{\frac{1}{m}}\zeta\cdot\big( \norm{ \vect{h}_{l-1}^{(j)}\cdot \matr{\Theta}_{l}^{(j)} - \vect{h}_{l-1}^{(0)}\cdot \matr{\Theta}_{l}^{(j)} }_{2}~+ \norm{ \vect{h}_{l-1}^{(0)}\cdot \matr{\Theta}_{l}^{(j)} - \vect{h}_{l-1}^{(0)}\cdot \matr{\Theta}_{l}^{(0)} }_{2}\big) \\
    & \leq \sqrt{\frac{1}{m}}\zeta\cdot( \norm{\matr{\Theta}_{l}^{(0)}}_{2} + \norm{\matr{\Theta}_{l}^{(j)} - \matr{\Theta}_{l}^{(0)}}_{F} ) \cdot\norm{\vect{h}_{l-1}^{(j)} - \vect{h}_{l-1}^{(0)}}_{2}~+\\
    &\qquad \sqrt{\frac{1}{m}}\zeta \cdot \norm{\vect{h}_{l-1}^{(0)}}_{2}\cdot \norm{\matr{\Theta}_{l}^{(j)} - \matr{\Theta}_{l}^{(0)}}_{F} \\
    & \leq \sqrt{\frac{1}{m}}\zeta\cdot( \beta_{L}\sqrt{m} + \Upsilon/\sqrt{m} ) \cdot\norm{\vect{h}_{l-1}^{(j)} - \vect{h}_{l-1}^{(0)}}_{2}~+ \zeta \cdot \beta'_{h}\cdot \Upsilon/m \\
    & \leq \zeta\cdot( \beta_{L} + \Upsilon/m ) \cdot \zeta \frac{\Upsilon}{m}\cdot\Lambda_{l-1}^{(j)} + \zeta \cdot \beta'_{h}\cdot \Upsilon/m \\
    &\leq \zeta \frac{\Upsilon}{m} \cdot(\beta_{h} + 2\zeta \beta_{L} \cdot \Lambda_{l-1}^{(j)})
    = \zeta \frac{\Upsilon}{m} \cdot \Lambda_{l}^{(j)}
\end{split}
\end{displaymath}
where the last two inequalities are derived by applying \textbf{Lemma}~\ref{lemma_after_GD_weight_matrices_bounds} and the hypothesis. For the aggregation module output $\vect{h}_{gnn}^{(0)}$,
\begin{displaymath}
\begin{split}
    &\norm{\vect{h}_{gnn}^{(j)} - \vect{h}_{gnn}^{(0)}}_{2} = \sqrt{\frac{1}{m}} \norm{\sigma(\matr{\Theta}_{gnn}^{(j)}\cdot \vect{h}_{S}) - \sigma(\matr{\Theta}_{gnn}^{(0)}\cdot \vect{h}_{S})}_{2} \\
    & \leq \frac{\zeta}{\sqrt{m}} \norm{\matr{\Theta}_{gnn}^{(j)} - \matr{\Theta}_{gnn}^{(0)}}_{F} \cdot \norm{\vect{h}_{S}}_{2} \leq \frac{\zeta\Upsilon}{m}\beta_{h}.
\end{split}
\end{displaymath}
Then, for the first layer $l=1$, we have
\begin{displaymath}
\begin{split}
    &\norm{\vect{h}_{1}^{(j)} - \vect{h}_{1}^{(0)}}_{2} = \sqrt{\frac{1}{m}} \norm{\sigma(\vect{x}'\cdot \matr{\Theta}_{1}^{(j)} ) - \sigma(\vect{x}'\cdot \matr{\Theta}_{1}^{(0)} )}_{2} \\
    & \leq \frac{\zeta}{\sqrt{m}} \norm{\matr{\Theta}_{gnn}^{(j)} - \matr{\Theta}_{1}^{(0)}}_{F} \cdot \norm{\vect{x}'}_{2} \leq \frac{\zeta\Upsilon}{m} \cdot (\zeta\cdot \beta_{L} + 1) \leq \frac{\zeta\Upsilon}{m}\cdot \beta_{h}.
\end{split}
\end{displaymath}
Combining all the results, for $\vect{h} \in \{\vect{h}_{gnn}, \vect{h}_{1}, \dots, \vect{h}_{L-1}\}$, it has
\begin{displaymath}
\begin{split}
    &\norm{\vect{h}^{(j)} - \vect{h}^{(0)}}_{2} \leq \frac{\zeta \Upsilon}{m}\cdot \beta_{h} \cdot \frac{(2\zeta \beta_{L})^{L}-1}{2\zeta \beta_{L}-1} = \Lambda^{(j)},
\end{split}
\end{displaymath}
which completes the proof.


\begin{lemma}
With initialized network parameters $\matr{\Theta}$ and the probability of at least $1 - O(TL)\cdot e^{-\Omega(m)}$, we have
\begin{displaymath}
\begin{split}
    &\norm{\nabla_{\matr{\Theta}}f(\widetilde{\matr{X}};\matr{\Theta}^{(0)})}_{F} \leq \beta_{h}\beta_{3}\cdot (\beta_{L}\zeta)^{L}/ m, \norm{\nabla_{\matr{\Theta}_{L}}f(\widetilde{\matr{X}};\matr{\Theta}^{(0)})}_{F} \leq \beta_{h}/\sqrt{m},
\end{split}
\end{displaymath}
and the norm of gradient difference
\begin{displaymath}
\begin{split}
    &\norm{\nabla_{\matr{\Theta}}f(\widetilde{\matr{X}};\matr{\Theta}^{(0)}) - \nabla_{\matr{\Theta}}f(\widetilde{\matr{X}};\matr{\Theta}^{(j)})}_{F} \leq 3\cdot \Lambda^{(j)}, \\
    &\norm{\nabla_{\matr{\Theta}_{L}}f(\widetilde{\matr{X}};\matr{\Theta}^{(0)}) - \nabla_{\matr{\Theta}_{L}}f(\widetilde{\matr{X}};\matr{\Theta}^{(j)})}_{F} \leq \Lambda^{(j)} / \sqrt{m}.
\end{split}
\end{displaymath}
with $\matr{\Theta} \in \{\matr{\Theta}_{gnn}, \matr{\Theta}_{1}, \dots, \matr{\Theta}_{L-1}\}$.

\label{lemma_after_GD_gradient_matrix_norm}
\end{lemma}


\textbf{Proof.}
First, for $l\in[L-1]$, we have
\begin{displaymath}
\begin{split}
    &\norm{\nabla_{\matr{\Theta}_{l}}f(\widetilde{\matr{X}};\matr{\Theta}^{(0)})}_{F} = 
    m^{-\frac{L-l+1}{2}}  
    \norm{\vect{h}_{l-1}^{(0)}  \big(\matr{\Theta}_{L}^{(0)}  
    \big(\prod_{q=l+1}^{L-1} \matr{\Gamma}_{q}  \cdot \matr{\Theta}_{q}^{(0)}\big)
    \cdot \matr{\Gamma}_{l} 
    \big)}_{F} \\
    & \leq m^{-\frac{L-l+1}{2}} \cdot \norm{\vect{h}_{l-1}^{(0)}\matr{\Theta}_{L}^{(0)}}_{F} \cdot \norm{\prod_{q=l}^{L-1} \matr{\Gamma}_{q}}_{2} \cdot \norm{\prod_{q=l+1}^{L-1} \matr{\Theta}_{q}^{(0)}}_{2} \\
    & \leq m^{-\frac{L-l+1}{2}} \cdot \norm{\vect{h}_{l-1}^{(0)}}_{2} \cdot \norm{\matr{\Theta}_{L}^{(0)}}_{2} \cdot \norm{\prod_{q=l}^{L-1} \matr{\Gamma}_{q}}_{2} \cdot \norm{\prod_{q=l+1}^{L-1} \matr{\Theta}_{q}^{(0)}}_{2} \\
    & \leq m^{-\frac{L-l+1}{2}}\cdot \beta_{h}\beta_{3} \cdot \zeta^{L-l} \cdot (\beta_{L} \sqrt{m})^{L-l-1}
    \leq \beta_{h}\beta_{3}\cdot (\beta_{L}\zeta)^{L}/ m.
\end{split} 
\end{displaymath}
For $\matr{\Theta}_{gnn}$, we can also derive similar results. For $\matr{\Theta}_{L}$,
\begin{displaymath}
\begin{split}
    &\norm{\nabla_{\matr{\Theta}_{L}}f(\widetilde{\matr{X}};\matr{\Theta}^{(0)})}_{F} = 
    m^{-0.5}\cdot \norm{\vect{h}_{L-1}^{(0)}}_{2} \leq \beta_{h} / \sqrt{m}
\end{split}
\end{displaymath}
Then, with $\nabla_{l}^{(j)} = m^{-\frac{L-l+1}{2}}\cdot (\matr{\Theta}_{L}^{(j)})^{\intercal} \cdot 
\big(\prod_{q=l+1}^{L-1} \matr{\Gamma}_{q}  \cdot (\matr{\Theta}_{q}^{(j)})^{\intercal}\big)
\cdot \matr{\Gamma}_{l}$, we have the norm of gradient difference 
\begin{displaymath}
\begin{split}
    &\norm{\nabla_{\matr{\Theta}_{l}}f(\widetilde{\matr{X}};\matr{\Theta}^{(j)}) - \nabla_{\matr{\Theta}_{l}}f(\widetilde{\matr{X}};\matr{\Theta}^{(0)})}_{F} 
    =  \norm{\vect{h}_{l-1}^{(0)} \cdot \nabla_{l}^{(0)} - \vect{h}_{l-1}^{(j)} \cdot \nabla_{l}^{(j)}}_{F} \\
    & \leq \norm{\vect{h}_{l-1}^{(0)} \cdot \nabla_{l}^{(0)} - \vect{h}_{l-1}^{(j)} \cdot \nabla_{l}^{(0)}}_{F} 
    + \norm{\vect{h}_{l-1}^{(j)} \cdot \nabla_{l}^{(0)} - \vect{h}_{l-1}^{(j)} \cdot \nabla_{l}^{(j)}}_{F} \\
    & \leq \norm{\vect{h}_{l-1}^{(j)}}_{F} \cdot \norm{\nabla_{l}^{(0)} - \nabla_{l}^{(j)}}_{F}
    + \norm{\nabla_{l}^{(0)}}_{F} \cdot \norm{\vect{h}_{l-1}^{(0)} - \vect{h}_{l-1}^{(j)}}_{F} \\
    & \leq \big(\beta_{h} + \Lambda^{(j)}\big)\cdot \norm{\nabla_{l}^{(0)} - \nabla_{l}^{(j)}}_{F} + \Lambda^{(j)}\cdot \norm{\nabla_{l}^{(0)}}_{F}.
\end{split} 
\end{displaymath}
Here, for the difference of $\nabla$, we have
\begin{displaymath}
\begin{split}
    &\norm{\nabla_{l}^{(0)} - \nabla_{l}^{(j)}}_{F} \\
    &= m^{-\frac{L-l+1}{2}} \norm{\matr{\Theta}_{L}^{(0)} \cdot 
    \big(\prod_{q=l+1}^{L-1} \matr{\Gamma}_{q}  \cdot \matr{\Theta}_{q}^{(0)}\big)
     \matr{\Gamma}_{l} 
    - \matr{\Theta}_{L}^{(j)}  
    \big(\prod_{q=l+1}^{L-1} \matr{\Gamma}_{q}   \matr{\Theta}_{q}^{(j)}\big)
     \matr{\Gamma}_{l}}_{F} \\
    & = m^{-\frac{1}{2}}\cdot \norm{
        \nabla_{l+1}^{(0)} \matr{\Gamma}_{l}  \cdot \matr{\Theta}_{l+1}^{(0)} - 
        \nabla_{l+1}^{(j)} \matr{\Gamma}_{l}  \cdot \matr{\Theta}_{l+1}^{(j)}
    }_{F} \\
    & \leq \frac{\zeta}{\sqrt{m}}\cdot( \norm{
        \nabla_{l+1}^{(0)} \cdot \matr{\Theta}_{l+1}^{(0)} -
        \nabla_{l+1}^{(0)} \cdot \matr{\Theta}_{l+1}^{(j)}
    }_{F} + 
    \norm{
        \nabla_{l+1}^{(0)} \cdot \matr{\Theta}_{l+1}^{(j)} - 
        \nabla_{l+1}^{(j)} \cdot \matr{\Theta}_{l+1}^{(j)}
    }_{F}) \\
    & \leq \frac{\zeta}{\sqrt{m}}\cdot (\norm{\nabla_{l+1}^{(0)}}_{F}\norm{
          \cdot \matr{\Theta}_{l+1}^{(0)} - \matr{\Theta}_{l+1}^{(j)}
    }_{F} + \norm{\matr{\Theta}_{l+1}^{(j)}}_{F}
    \norm{
        \nabla_{l+1}^{(0)} - \nabla_{l+1}^{(j)}
    }_{F}).
\end{split} 
\end{displaymath}
To continue the proof, we need to bound the term $\norm{\nabla_{l}^{(0)}}_{F}$ as
\begin{displaymath}
\begin{split}
    \norm{\nabla_{l}^{(0)}}_{F} = m^{-0.5} \norm{\matr{\Gamma}_{l}\matr{\Theta}_{l+1}^{(0)}\cdot\nabla_{l+1}^{(0)}}_{F} \leq \zeta \beta_{L}\cdot \norm{\nabla_{l+1}^{(0)}}_{F}.
\end{split} 
\end{displaymath}
Since for $l=L-1$ we have
\begin{displaymath}
\begin{split}
    \norm{\nabla_{L - 1}^{(0)}}_{F} \leq \frac{\zeta \beta_{3}}{m},
\end{split} 
\end{displaymath}
we can derive
\begin{displaymath}
\begin{split}
    \norm{\nabla_{l}^{(0)}}_{F} \leq \frac{\beta_{3}}{m}\cdot (\zeta\cdot \beta_{L})^{L} \leq 1
\end{split} 
\end{displaymath}
with sufficiently large $m$, and this bound also applies to $\norm{\nabla_{gnn}^{(0)}}_{F}$. For 
\begin{displaymath}
\begin{split}
    &\norm{\nabla_{L}^{(0)} - \nabla_{L}^{(j)}}_{F} = m^{-0.5} \norm{\vect{h}_{L-1}^{(0)} - \vect{h}_{L-1}^{(j)}}_{F} \leq \Lambda^{(j)} / \sqrt{m}
\end{split} 
\end{displaymath}
Therefore, we have
\begin{displaymath}
\begin{split}
    &\norm{\nabla_{l}^{(0)} - \nabla_{l}^{(j)}}_{F}
    \leq \frac{\zeta\Upsilon}{m}
    + \zeta\cdot (\beta_{L} + \Upsilon / m)
    \norm{
        \nabla_{l+1}^{(0)} - \nabla_{l+1}^{(j)}
    }_{F}.
\end{split} 
\end{displaymath}
By following a similar approach as in Lemma~\ref{lemma_after_GD_model_results_variables}, we will have
\begin{displaymath}
\begin{split}
    &\norm{\nabla_{l}^{(0)} - \nabla_{l}^{(j)}}_{F} 
    \leq \frac{\zeta \Upsilon}{m} \cdot \frac{(2\zeta \beta_{L})^{L}-1}{2\zeta \beta_{L}-1} = \frac{\Lambda^{(j)}}{\beta_{h}}.
\end{split} 
\end{displaymath}
Therefore, we will have
\begin{displaymath}
\begin{split}
    \norm{\nabla_{\matr{\Theta}_{l}}f(\widetilde{\matr{X}};\matr{\Theta}^{(j)}) &- \nabla_{\matr{\Theta}_{l}}f(\widetilde{\matr{X}};\matr{\Theta}^{(0)})}_{F}  
    \leq \big(\beta_{h} + \Lambda^{(j)}\big)\cdot \frac{\Lambda^{(j)}}{\beta_{h}} + \Lambda^{(j)} \\
    &\leq \frac{\Lambda^{(j)}}{\beta_{h}}\cdot (2\beta_{h} + 1) = \Lambda^{(j)}\cdot (2 + \frac{1}{\beta_{h}}) \leq 3\cdot \Lambda^{(j)}
\end{split} 
\end{displaymath}
with sufficiently large $m$. This bound can also be derived for $\norm{\nabla_{\matr{\Theta}_{gnn}}f(\widetilde{\matr{X}};\matr{\Theta}^{(0)})}_{2}$ with a similar procedure.
For $L$-th layer, we have
\begin{displaymath}
\begin{split}
    \norm{\nabla_{\matr{\Theta}_{L}}f(\widetilde{\matr{X}};\matr{\Theta}^{(j)}) &- \nabla_{\matr{\Theta}_{L}}f(\widetilde{\matr{X}};\matr{\Theta}^{(0)})}_{F}  
    \leq m^{-0.5}\cdot \norm{\vect{h}_{L-1}^{(0)} - \vect{h}_{L-1}^{(j)}}_{F} \\
    & \leq \Lambda^{(j)} / \sqrt{m},
\end{split} 
\end{displaymath}
which completes the proof.

\begin{lemma}
With the probability of at least $1 - O(TL)\cdot e^{-\Omega(m)}$, we have the gradient for all the network as
\begin{displaymath}
\begin{split}
    &\norm{g(\widetilde{\matr{X}};\matr{\Theta}^{(0)})}_{2} \leq m^{-1}\beta_{h} \cdot \sqrt{L\cdot \beta_{3}^{2}\cdot (\beta_{L}\zeta)^{2L} + m}, \\
    &\norm{g(\widetilde{\matr{X}};\matr{\Theta}^{(j)})}_{2} \leq \Lambda^{(j)}\cdot\sqrt{9L + m^{-1}} + m^{-1}\beta_{h} \cdot \sqrt{L\cdot \beta_{3}^{2}\cdot (\beta_{L}\zeta)^{2L} + m} \\
    &\norm{g(\widetilde{\matr{X}};\matr{\Theta}^{(0)}) - g(\widetilde{\matr{X}};\matr{\Theta}^{(j)})}_{2} \leq \Lambda^{(j)}\cdot\sqrt{9L + m^{-1}}.
\end{split}
\end{displaymath}
\label{lemma_after_GD_gradient_for_network_norm}
\end{lemma}


\textbf{Proof.}
First, for the gradient before GD, we have
\begin{displaymath}
\begin{split}
    &\norm{g(\widetilde{\matr{X}};\matr{\Theta}^{(0)})}_{2} = \sqrt{
    \norm{\nabla_{\matr{\Theta}_{gnn}}f(\widetilde{\matr{X}};\matr{\Theta}^{(0)})}_{2}^{2}
    + \sum_{l=1}^{L} \norm{\nabla_{\matr{\Theta}_{l}}f(\widetilde{\matr{X}};\matr{\Theta}^{(0)})}_{2}^{2}
    } \\
    & \leq m^{-1}\beta_{h} \cdot \sqrt{L\cdot \beta_{3}^{2}\cdot (\beta_{L}\zeta)^{2L} + m}.
\end{split}
\end{displaymath}
Then, for the norm of gradients, $\matr{\Theta} \in \{\matr{\Theta}_{gnn}, \matr{\Theta}_{1}, \dots, \matr{\Theta}_{L-1}\}$, we have
\begin{displaymath}
\begin{split}
    &\norm{g(\widetilde{\matr{X}};\matr{\Theta}^{(0)}) - g(\widetilde{\matr{X}};\matr{\Theta}^{(j)})}_{2} \\
    & = \sqrt{
    \sum_{\matr{\Theta}} \norm{\nabla_{\matr{\Theta}}f(\widetilde{\matr{X}};\matr{\Theta}^{(0)}) - \nabla_{\matr{\Theta}}f(\widetilde{\matr{X}};\matr{\Theta}^{(j)})}_{2}^{2}
    } \\
    & \leq \sqrt{9L\cdot (\Lambda^{(j)})^{2} + (\Lambda^{(j)})^{2} / m}  = \Lambda^{(j)}\cdot\sqrt{9L + m^{-1}}.
\end{split}
\end{displaymath}
Then, for the network gradient after GD, we have
\begin{displaymath}
\begin{split}
    &\norm{g(\widetilde{\matr{X}};\matr{\Theta}^{(j)})}_{2} \leq \norm{g(\widetilde{\matr{X}};\matr{\Theta}^{(0)}) - g(\widetilde{\matr{X}};\matr{\Theta}^{(j)})}_{2} + \norm{g(\widetilde{\matr{X}};\matr{\Theta}^{(0)})}_{2} \\
    & \leq \Lambda^{(j)}\cdot\sqrt{9L + m^{-1}} + m^{-1}\beta_{h} \cdot \sqrt{L\cdot \beta_{3}^{2}\cdot (\beta_{L}\zeta)^{2L} + m}
\end{split}
\end{displaymath}

\begin{lemma}
With the probability of at least $1 - O(TL)\cdot e^{-\Omega(m)}$, for the initialized parameter $\matr{\Theta}^{(0)}$, we have 
\begin{displaymath}
\begin{split}
    &\abs{f(\widetilde{\matr{X}};\matr{\Theta}^{(j)}) - \inp{g(\widetilde{\matr{X}};\matr{\Theta}^{(0)})}{\matr{\Theta}^{(j)} - \matr{\Theta}^{(0)}}} \\
    &\qquad\qquad \leq m^{-0.5} \cdot \big( \Lambda^{(j)} (1 + \beta_{3}) + \beta_{3}\beta_{h} + L\cdot \beta_{h} \Upsilon \big),
\end{split}
\end{displaymath}
and for the network parameter after GD, $\matr{\Theta}^{(j)}$, we have
\begin{displaymath}
\begin{split}
    \abs{f&(\widetilde{\matr{X}};\matr{\Theta}^{(j)}) - \inp{g(\widetilde{\matr{X}};\matr{\Theta}^{(j)})}{\matr{\Theta}^{(j)} - \matr{\Theta}^{(0)}}} \leq B_{3} \\
    & = m^{-0.5} \big(
        \beta_{3} (\Lambda^{(j)}+\beta_{h}) + L\cdot\Upsilon\cdot (\Lambda^{(j)}+\beta_{h})(\Lambda^{(j)} / \beta_{h} + 1) \big).
\end{split}
\end{displaymath}
\label{lemma_output_minus_inner_product}
\end{lemma}


\textbf{Proof.}
For the sake of enumeration, we let $\matr{\Theta}_{0} = \matr{\Theta}_{gnn}, \nabla_{0} = \nabla_{gnn}, \vect{h}_{0} = \vect{h}_{gnn}$ and $\vect{h}_{-1} = \vect{h}_{S}$. Then, we can derive
\begin{displaymath}
\begin{split}
    &\abs{f(\widetilde{\matr{X}};\matr{\Theta}^{(j)}) - \inp{g(\widetilde{\matr{X}};\matr{\Theta}^{(0)})}{\matr{\Theta}^{(j)} - \matr{\Theta}^{(0)}}}  = \abs{\frac{1}{\sqrt{m}}\inp{\vect{h}_{L-1}^{(j)}}{\matr{\Theta}_{L}^{(j)}} \\
    & \qquad - \frac{1}{\sqrt{m}} \inp{\vect{h}_{L-1}^{(0)}}{\matr{\Theta}_{L}^{(0)} - \matr{\Theta}_{L}^{(j)}} - \sum_{l=0}^{L-1} (\vect{h}_{l-1}^{(0)})^{\intercal} (\matr{\Theta}_{l}^{(0)} - \matr{\Theta}_{l}^{(j)})\nabla_{l}^{(0)}) } \\
    & \leq m^{-0.5}\norm{\vect{h}_{L}^{(j)} - \vect{h}_{L}^{(0)}}_{2} \norm{\matr{\Theta}_{L}^{(j)}}_{2}
    + m^{-0.5} \norm{\vect{h}_{L-1}^{(0)}}_{2} \norm{\matr{\Theta}_{L}^{(0)}}_{2} \\
    & \qquad\qquad + \sum_{l=0}^{L-1} \norm{\vect{h}_{l-1}^{(0)}}_{2} \norm{\matr{\Theta}_{l}^{(0)} - \matr{\Theta}_{l}^{(j)}}_{F} \norm{\nabla_{l}^{(0)}}_{F} \\
    & \leq m^{-0.5}\Lambda^{(j)} (\Upsilon / \sqrt{m} + \beta_{3}) + m^{-0.5} \beta_{3} \beta_{h} 
    + L\cdot \beta_{h} \frac{\Upsilon}{\sqrt{m}} \\
    & \leq m^{-0.5} \cdot \big( \Lambda^{(j)} (1 + \beta_{3}) + \beta_{3}\beta_{h} + L\cdot \beta_{h} \Upsilon \big).
\end{split}
\end{displaymath}
On the other hand, for network parameter after GD, we can have
\begin{displaymath}
\begin{split}
    &\abs{f(\widetilde{\matr{X}};\matr{\Theta}^{(j)}) - \inp{g(\widetilde{\matr{X}};\matr{\Theta}^{(j)})}{\matr{\Theta}^{(j)} - \matr{\Theta}^{(0)}}}  = \abs{\frac{1}{\sqrt{m}}\inp{\vect{h}_{L-1}^{(j)}}{\matr{\Theta}_{L}^{(j)}} \\
    & \qquad - \frac{1}{\sqrt{m}} \inp{\vect{h}_{L-1}^{(j)}}{\matr{\Theta}_{L}^{(j)} - \matr{\Theta}_{L}^{(0)}} - \sum_{l=0}^{L-1} (\vect{h}_{l-1}^{(j)})^{\intercal} (\matr{\Theta}_{l}^{(0)} - \matr{\Theta}_{l}^{(j)})\nabla_{l}^{(j)}) } \\
    & \leq \abs{m^{-0.5} \inp{\vect{h}_{L-1}^{(j)}}{\matr{\Theta}_{L}^{(0)}} -
    \sum_{l=0}^{L-1} (\vect{h}_{l-1}^{(j)})^{\intercal} (\matr{\Theta}_{l}^{(0)} - \matr{\Theta}_{l}^{(j)})\nabla_{l}^{(j)}} \\
    & \leq m^{-0.5} \norm{\vect{h}_{L-1}^{(j)}}_{2}\norm{\matr{\Theta}_{L}^{(0)}}_{2}
    + \sum_{l=0}^{L-1} \norm{\vect{h}_{l-1}^{(j)}}_{2} \norm{\matr{\Theta}_{l}^{(0)} - \matr{\Theta}_{l}^{(j)}}_{F} \norm{\nabla_{l}^{(j)}}_{F} \\
    & \leq m^{-0.5} \beta_{3} (\Lambda^{(j)}+\beta_{h}) + L\cdot (\Lambda^{(j)}+\beta_{h}) (\Upsilon / \sqrt{m}) (\Lambda^{(j)} / \beta_{h} + 1) \\ 
    & \leq m^{-0.5} \big(
        \beta_{3} (\Lambda^{(j)}+\beta_{h}) + L\cdot\Upsilon\cdot (\Lambda^{(j)}+\beta_{h})(\Lambda^{(j)} / \beta_{h} + 1) 
    \big).
\end{split}
\end{displaymath}
This completes the proof.

\textbf{Proof sketch for Lemmas \ref{lemma_initial_param_outputs}-\ref{lemma_output_minus_inner_product}.}
First we derive the conclusions in Lemma \ref{lemma_initial_param_outputs} with the property of Gaussian matrices. Then, Lemmas \ref{lemma_after_GD_weight_matrices_bounds} and \ref{lemma_after_GD_model_results_variables} are proved through the induction after breaking the target into norms of individual terms (variables, weight matrices) and applying Lemma \ref{lemma_initial_param_outputs}. Finally, for Lemmas \ref{lemma_after_GD_gradient_matrix_norm}-\ref{lemma_output_minus_inner_product}, we also decompose targets into norms of individual terms. Then, applying Lemmas \ref{lemma_initial_param_outputs}-\ref{lemma_after_GD_model_results_variables} the to bound these terms (at random initialization / after GD) would give the result.        $\blacksquare$

\section{Lemmas for Gradient Matrices}
Inspired by \cite{Neural-UCB,CNN_UCB-ban2021convolutional} and with sufficiently large network width $m$, the trained network parameter can be related to ridge regression estimator where the context is embedded by network gradients.  
With the received contexts and rewards up to time step $t$, we have the estimated parameter $\widehat{\matr{\Theta}}$ as
$
    \widehat{\matr{\Theta}}_{0} = (\matr{Z}_{0})^{-1}\cdot \vect{b}_{0}
$
where 
$
    \matr{Z}_{0} = \lambda\matr{I} + \frac{1}{m}\sum_{\tau=1}^{t} g(\widetilde{\matr{X}}_{\tau}; \matr{\Theta}_{0})g(\widetilde{\matr{X}}_{\tau}; \matr{\Theta}_{0})^{\intercal},  \vect{b}_{0}  = \frac{1}{\sqrt{m}}\sum_{\tau=1}^{t} r_{\tau} \cdot g(\widetilde{\matr{X}}_{\tau}; \matr{\Theta}_{0}).
$
We also define the gradient matrix w.r.t. the network parameters as
\begin{displaymath}
\begin{split}
    & \matr{G}^{(j)} = \big( g(\widetilde{\matr{X}}_{1}; \matr{\Theta}^{(j)}), \dots, g(\widetilde{\matr{X}}_{t}; \matr{\Theta}^{(j)}) \big) \\
    & \vect{f}^{(j)} = \big( f(\widetilde{\matr{X}}_{1}; \matr{\Theta}^{(j)}), \dots, f(\widetilde{\matr{X}}_{t}; \matr{\Theta}^{(j)}) \big), 
    \quad \vect{r} = \big( r_{1}, \dots, r_{t} \big) \\
    & \matr{\Theta}^{(j+1)} = \matr{\Theta}^{(j)} - \eta\cdot \big( (\matr{G}^{(j)})^{\intercal} (\vect{f}^{(j)} - \vect{r}) \big).
\end{split}
\end{displaymath}
where the $t$ notation is omitted by default. Then, we use the following Lemma to bound the above matrices.

\begin{lemma}
After $j$ iterations, with the probability of at least $1 - O(L)\cdot e^{-\Omega(m)}$, we have 
\begin{displaymath}
\begin{split}
    & \norm{\matr{G}^{(0)}}_{F} \leq G_{1} = m^{-1}\beta_{h} \cdot \sqrt{t\cdot (L\cdot \beta_{3}^{2}\cdot (\beta_{L}\zeta)^{2L} + m)}, \\
    & \norm{\matr{G}^{(0)} - \matr{G}^{(j)}}_{F} \leq \Lambda^{(j)}\cdot\sqrt{t\cdot(9L + m^{-1})}, \\
    & \norm{\matr{G}^{(j)}}_{F} \leq \widetilde{I}_{1} = \sqrt{t\cdot (L\cdot \beta_{3}^{2}\cdot (\beta_{L}\zeta)^{2L} + m)} + \Lambda^{(j)}\sqrt{t\cdot(9L + m^{-1})}\\
    & \norm{\vect{f}^{(j)} - (\matr{G}^{(j)})^{\intercal}(\widehat{\matr{\Theta}}^{(j)} - \widehat{\matr{\Theta}}^{(0)})}_{2} \leq \sqrt{t}\cdot B_{3} \\
    & ~~ = \sqrt{t}\cdot  m^{-0.5} \big(
        \beta_{3} (\Lambda^{(j)}+\beta_{h}) + L\cdot\Upsilon\cdot (\Lambda^{(j)}+\beta_{h})(\Lambda^{(j)} / \beta_{h} + 1) \big)
\end{split}
\end{displaymath}

\label{lemma_bounding_gradient_matrix_at_init}
\end{lemma}


\textbf{Proof.}
For the gradient matrix after random initialization, we have
\begin{displaymath}
\begin{split}
    & \norm{\matr{G}^{(0)}}_{F} = \sqrt{\sum_{\tau=1}^{t} \norm{g(\widetilde{\matr{X}}_{\tau}; \matr{\Theta}^{(0)})}_{2}^{2}}
    \leq m^{-1}\beta_{h} \cdot \sqrt{t\cdot L\cdot \beta_{3}^{2}\cdot (\beta_{L}\zeta)^{2L} + m}
\end{split}
\end{displaymath}
with the conclusion from Lemma \ref{lemma_after_GD_gradient_for_network_norm}. Then, 
\begin{displaymath}
\begin{split}
    & \norm{\matr{G}^{(0)} - \matr{G}^{(j)}}_{F} = \sqrt{\sum_{\tau=1}^{t} \norm{g(\widetilde{\matr{X}}_{\tau}; \matr{\Theta}^{(0)}) - g(\widetilde{\matr{X}}_{\tau}; \matr{\Theta}^{(j)})}_{2}^{2}} \\
    & \qquad\qquad \leq \Lambda^{(j)}\cdot\sqrt{t\cdot(9L + m^{-1})}.
\end{split}
\end{displaymath}
For the third inequality in this Lemma, we have
\begin{displaymath}
\begin{split}
    & \norm{\vect{f}^{(j)} - (\matr{G}^{(j)})^{\intercal}(\widehat{\matr{\Theta}}^{(j)} - \widehat{\matr{\Theta}}^{(0)})}_{2} \\
    & = \sqrt{\sum_{\tau=1}^{t} \abs{f(\widetilde{\matr{X}}_{\tau}; \matr{\Theta}^{(j)}) - 
    \inp{g(\widetilde{\matr{X}}_{\tau};\matr{\Theta}^{(j)})}{\matr{\Theta}^{(j)} - \matr{\Theta}^{(0)}})
    }^{2}} \\
    & \leq \sqrt{t}\cdot  m^{-0.5} \big(
        \beta_{3} (\Lambda^{(j)}+\beta_{h}) + L\cdot\Upsilon\cdot (\Lambda^{(j)}+\beta_{h})(\Lambda^{(j)} / \beta_{h} + 1) \big)
\end{split}
\end{displaymath}
based on Lemma \ref{lemma_output_minus_inner_product}.

Analogous to \cite{CNN_UCB-ban2021convolutional,Neural-UCB}, we define another auxiliary sequence to bound the parameter difference. With $\widetilde{\matr{\Theta}}^{(0)} = \matr{\Theta}^{(0)}$, we have
$
    \widetilde{\matr{\Theta}}^{(j+1)} = \\
     \widetilde{\matr{\Theta}}^{(j)} - 
     \eta\cdot \bigg( \matr{G}^{(j)} \big((\matr{G}^{(j)})^{\intercal}(\widetilde{\matr{\Theta}}^{(j)} - \widetilde{\matr{\Theta}}^{(0)}) - \vect{r}\big) + m\lambda (\widetilde{\matr{\Theta}}^{(j)} - \widetilde{\matr{\Theta}}^{(0)})\bigg)
$. 
\begin{lemma}
After $j$ iterations, with the probability of at least $1 - O(L)\cdot e^{-\Omega(m)}$, we have 
\begin{displaymath}
\begin{split}
    & \norm{\widetilde{\matr{\Theta}}^{(j)} - \matr{\Theta}^{(0)} - \widehat{\matr{\Theta}}_{t} / \sqrt{m}}_{2}
    \leq \sqrt{t / (m\lambda)}
\end{split}
\end{displaymath}

\label{lemma_bounding_auxiliary_parameter}
\end{lemma}

\textbf{Proof.}
The proof is analogous to Lemma 10.2 in \cite{CNN_UCB-ban2021convolutional} and Lemma C.4 in \cite{Neural-UCB}. Switching $\matr{G}_{0}$ to $\matr{G}_{j}$ would give the result.   $\blacksquare$

Then, we can have the following lemma to bridge the difference between the regression estimator $\widehat{\matr{\Theta}}$ and the network parameter $\matr{\Theta}$.

\begin{lemma}
At this time step $t$, with the notation defined in Lemma \ref{lemma_CB_one_step_same_graph} and the probability at least $1 - O(L)\cdot e^{-\Omega(m)}$, we will have
\begin{displaymath}
\begin{split}
    & \norm{\matr{\Theta}_{t} - \matr{\Theta}_{0} - \widehat{\matr{\Theta}}_{t} / \sqrt{m}}_{2} \leq (\widetilde{I}_{1} \cdot \sqrt{t} B_{3} +  m\cdot \widetilde{I}_{2}) / (m\lambda) + \sqrt{t / (m\lambda)}
\end{split}
\end{displaymath}
with proper $m, \eta$ as in Lemma \ref{lemma_CB_one_step_same_graph}
\label{lemma_linking_regre_est_with_net_param}
\end{lemma}


\textbf{Proof.}
With an analogous approach from Lemma 6.2 in \cite{CNN_UCB-ban2021convolutional}, we can have
\begin{displaymath}
\begin{split}
    & \norm{\widetilde{\matr{\Theta}}^{(j+1)} - \matr{\Theta}^{(j+1)}}_{2} \\
    & \leq
    \eta \norm{\matr{G}^{(j)}}_{2} \norm{\vect{f}^{(j)}  - (\matr{G}^{(j)})^{\intercal} (\matr{\Theta}^{(j)} - \matr{\Theta}^{(0)})}_{2} + \eta m\lambda\norm{\matr{\Theta}^{(0)} - \matr{\Theta}^{(j)}}_{2} \\
    & + \norm{\matr{I} - \eta\cdot (m\lambda \matr{I} + \matr{G}^{(j)}(\matr{G}^{(j)})^{\intercal})}_{2}
    \norm{\matr{\Theta}^{(j)} - \widetilde{\matr{\Theta}}^{(j)}}_{2} = I_{1} + I_{2} + I_{3}.
\end{split}
\end{displaymath}
With Lemma \ref{lemma_bounding_gradient_matrix_at_init}, we can bound them as
\begin{displaymath}
\begin{split}
    & I_{1} \leq \eta \cdot \widetilde{I}_{1} \cdot \sqrt{t} B_{3}  \\
    & I_{2} \leq \eta m\lambda \sqrt{\sum_{i=0}^{L} \norm{\matr{\Theta}_{l}^{(0)} - \matr{\Theta}_{l}^{(j)}}_{F}^{2}} \leq \eta m\cdot \widetilde{I}_{2} =\eta m \lambda \sqrt{L+1} \cdot \Upsilon / \sqrt{m} .
\end{split}
\end{displaymath}
based on the conclusion from Lemma \ref{lemma_after_GD_weight_matrices_bounds}. 
For $I_{3}$, we have
\begin{displaymath}
\begin{split}
    & \eta\cdot (m\lambda \matr{I} + \matr{G}^{(0)}(\matr{G}^{(0)})^{\intercal}) \preceq 
    \eta\cdot \matr{I}\\
    & \qquad\qquad \cdot \big(m\lambda + (m^{-1}\beta_{h} \cdot \sqrt{t\cdot (L\cdot \beta_{3}^{2}\cdot (\beta_{L}\zeta)^{2L} + m)})^{2}\big) 
    \preceq \matr{I}
\end{split}
\end{displaymath}
with proper choice of $m$ and $\eta$.
It leads to 
\begin{displaymath}
\begin{split}
    & \norm{\widetilde{\matr{\Theta}}^{(j+1)} - \matr{\Theta}^{(j+1)}}_{2} 
    \leq (1 - \eta m \lambda) \norm{\widetilde{\matr{\Theta}}^{(j)} - \matr{\Theta}^{(j)}}_{2} + \widetilde{I}_{1} \cdot \sqrt{t} B_{3} + \eta m\cdot \widetilde{I}_{2}
\end{split}
\end{displaymath}
which by induction and $\widetilde{\matr{\Theta}}^{(0)}=\matr{\Theta}^{(0)}$, we have
\begin{displaymath}
\begin{split}
    & \norm{\widetilde{\matr{\Theta}}^{(j)} - \matr{\Theta}^{(j)}}_{2} 
    \leq (\widetilde{I}_{1} \cdot \sqrt{t} B_{3} +  m\cdot \widetilde{I}_{2}) / (m\lambda).
\end{split}
\end{displaymath}
Finally, 
\begin{displaymath}
\begin{split}
    & \norm{\matr{\Theta}_{t} - \matr{\Theta}_{0} - \widehat{\matr{\Theta}}_{t} / \sqrt{m}}_{2} \leq \norm{\widetilde{\matr{\Theta}}^{(j)} - \matr{\Theta}^{(j)}}_{2} + 
    \norm{\widetilde{\matr{\Theta}}_{t} - \matr{\Theta}_{0} - \widehat{\matr{\Theta}}_{0} / \sqrt{m}}_{2} \\
    & \leq (\widetilde{I}_{1} \cdot \sqrt{t} B_{3} +  m\cdot \widetilde{I}_{2}) / (m\lambda) + \sqrt{t / (m\lambda)},
\end{split}
\end{displaymath}
which completes the proof. 

\begin{lemma}
At this time step $t$, with the probability at least $1 - O(L)\cdot e^{-\Omega(m)}$, we will have
\begin{displaymath}
\begin{split}
    & \norm{\matr{Z}_{t}}_{2} \leq \lambda + \\
    & \quad \frac{t(L+1)}{m}\big(\Lambda^{(j)}\cdot\sqrt{9L + m^{-1}} + m^{-1}\beta_{h} \cdot \sqrt{L\cdot \beta_{3}^{2}\cdot (\beta_{L}\zeta)^{2L} + m}\big)^{2}, \\
    & \norm{\matr{G}_{t}^{\intercal}\matr{G}_{t} - \matr{G}_{0}^{\intercal}\matr{G}_{0}}_{F} \leq 
    2 t\cdot m^{-1}(\Lambda^{(j)}\cdot\sqrt{9L + m^{-1}}) \\
    &\quad \cdot \big(\Lambda^{(j)}\sqrt{9L + m^{-1}} + m^{-1}\beta_{h} \cdot \sqrt{L\cdot \beta_{3}^{2}\cdot (\beta_{L}\zeta)^{2L} + m} \big) = B_{G} / m
\end{split}
\end{displaymath}
with proper $m, \eta$ as in Lemma \ref{lemma_CB_one_step_same_graph}.
\label{lemma_bound_G_t_A_t}
\end{lemma}


\textbf{Proof.}
For the gradient matrix of ridge regression, we have
\begin{displaymath}
\begin{split}
    & \norm{\matr{Z}_{t}}_{2} 
    \leq \lambda + m^{-1}\sum_{\tau=1}^{t} \norm{g(\widetilde{\matr{X}}_{\tau}; \matr{\Theta}_{t})}_{2}^{2} \leq \lambda + \\
    & \quad \frac{t(L+1)}{m}\big(\Lambda^{(j)}\cdot\sqrt{9L + m^{-1}} + m^{-1}\beta_{h} \cdot \sqrt{L\cdot \beta_{3}^{2}\cdot (\beta_{L}\zeta)^{2L} + m}\big)^{2}
\end{split}
\end{displaymath}
with the results from Lemma \ref{lemma_after_GD_gradient_for_network_norm}. Then,
\begin{displaymath}
\begin{split}
    & \norm{\matr{G}_{t}^{\intercal}\matr{G}_{t} - \matr{G}_{0}^{\intercal}\matr{G}_{0}}_{F} 
    \leq m^{-1} \cdot \\
    & \sqrt{
    \sum_{i, j = 1}^{t} 
    \norm{g(\widetilde{\matr{X}}_{i}; \matr{\Theta}_{t}) + g(\widetilde{\matr{X}}_{j}; \matr{\Theta}_{0})}_{2}^{2} + 
    \norm{g(\widetilde{\matr{X}}_{i}; \matr{\Theta}_{t}) - g(\widetilde{\matr{X}}_{j}; \matr{\Theta}_{0})}_{2}^{2}
    } \\
    & \leq 2 t\cdot m^{-1} \cdot \big(\Lambda^{(j)}\sqrt{9L + m^{-1}} + m^{-1}\beta_{h} \cdot \sqrt{L\cdot \beta_{3}^{2}\cdot (\beta_{L}\zeta)^{2L} + m} \big) \\
    & \qquad (\Lambda^{(j)}\cdot\sqrt{9L + m^{-1}}) = B_{G} / m.
\end{split}
\end{displaymath}
The proof is then completed. 


\textbf{Proof sketch for Lemmas \ref{lemma_bounding_gradient_matrix_at_init}-\ref{lemma_bound_G_t_A_t}.}
Analogous to lemmas in Section \ref{sec_appendix_A}, Lemma \ref{lemma_bounding_gradient_matrix_at_init} is proved by Lemmas \ref{lemma_after_GD_gradient_for_network_norm}, \ref{lemma_output_minus_inner_product} by breaking the target into the product of norms. The proof of Lemma \ref{lemma_bounding_auxiliary_parameter} is analogous to Lemma 10.2 in \cite{CNN_UCB-ban2021convolutional} and Lemma C.4 in \cite{Neural-UCB}, then replacing $\matr{G}_{0}$ with $\matr{G}_{j}$ would give the result. Then, based on Lemma \ref{lemma_bounding_auxiliary_parameter} results, Lemma \ref{lemma_linking_regre_est_with_net_param} will be proved with after bounding $\norm{\widetilde{\matr{\Theta}}^{(j+1)} - \matr{\Theta}^{(j+1)}}_{2}$ by induction. Finally, 
Lemma \ref{lemma_bound_G_t_A_t} is proved by decomposing the norm into sum of individual terms, and bounding these terms with bounds on gradients in Lemma \ref{lemma_after_GD_gradient_for_network_norm}.
$\blacksquare$


\section{Lemmas for Model Convergence}
\begin{lemma}
After $T$ time steps, assume the model with width $m$ defined in Lemma \ref{lemma_CB_one_step_same_graph} are trained with the $J$-iterations GD on the past contexts and rewards. Then, there exists a constant $\beta_{F}$, such that $\beta_{F}\cdot \eta < 1$, for any $j\in [J]$:
\begin{displaymath}
\begin{split}
    \norm{\matr{V}^{(j)}}_{2} \leq \frac{1}{4}\eta \beta_{F} \cdot \norm{\vect{F}^{(j)}_{T} - \vect{Y}_{T}}_{2}
\end{split}
\end{displaymath}
where $\vect{F}^{(j)} = [f(\mathcal{G}_{T}, \widetilde{\matr{X}}_{\tau};\matr{\Theta}^{(j)})]_{\tau=1}^{T}$, and $\vect{Y}_{T} = [r_{\tau}]_{\tau=1}^{T}$.

\label{lemma_after_GD_I_2_term}
\end{lemma}

\textbf{Proof.}
We prove this lemma following an analogous approach as Lemma B.6 in \cite{skip_kernel_du2019gradient}.
Given $\widetilde{\matr{X}}$, we denote 
$\nabla\mathcal{L}({\matr{\Theta}}^{(j)}) = \frac{\partial~\mathcal{L}(\matr{\Theta}^{(j)})}{\partial~\matr{\matr{\Theta}}}$, and 
$\nabla f({\matr{\Theta}}^{(j)}) = \frac{\partial~f(\mathcal{G}_{T}, \widetilde{\matr{X}}; \matr{\Theta}^{(j)})}{\partial~\matr{\Theta}}$, where $\matr{\Theta}\in\{\matr{\Theta}_{gnn}, \matr{\Theta}_{1}, \dots, \matr{\Theta}_{L}\}$. 
By the definition of $\norm{\matr{V}^{(j)}}$, we have its element $\abs{\matr{V}^{(j)}(\widetilde{\matr{X}})}$
\begin{displaymath}
\begin{split}
    &  \leq \eta \cdot \max_{0\leq s\leq \eta} \bigg[ \sum_{\Theta} 
    \norm{\nabla\mathcal{L}({\Theta}^{(j)})}_{F} 
    \norm{\nabla f({\Theta}^{(j)}) - \nabla f({\Theta}^{(j)}, s)}_{F}
    \bigg].
\end{split}
\end{displaymath}

%
With the notation and conclusion from Lemma \ref{lemma_after_GD_weight_matrices_bounds}, we have
\begin{displaymath}
\begin{split}
    & \norm{\nabla\mathcal{L}({\Theta}^{(j)})}_{F} 
    \leq m^{-\frac{1}{2}} 2\sqrt{T} \norm{\vect{F}^{(j)}_{T}- \vect{Y}_{T}}_{2} \cdot \zeta^{L} \cdot(2\beta_{L})^{L} \beta_{h}
\end{split}
\end{displaymath}

Meantime,
$
    \norm{\nabla f({\Theta}_{l}^{(j)}) - \nabla f({\Theta}_{l}^{(j)}, s)}_{F} = m^{-\frac{L-l+1}{2}} \\ 
    \norm{ \vect{h}^{(j)}_{l-1} (\matr{\Theta}^{(j)}_{L})^{\intercal}
    \big(\prod_{q=l+1}^{L-1} \matr{\Gamma}^{(j)}_{q}  (\matr{\Theta}^{(j)}_{q})^{\intercal} \big) 
    \cdot \matr{\Gamma}^{(j)}_{l} 
    - 
    \vect{h}^{(j), s}_{l-1} (\matr{\Theta}^{(j), s}_{L})^{\intercal}  \\
    \big(\prod_{q=l+1}^{L-1} \matr{\Gamma}^{(j), s}_{q}  \cdot (\matr{\Theta}^{(j), s}_{q})^{\intercal} \big)
    \cdot \matr{\Gamma}^{(j), s}_{l} }_{F}.
$
A similar form can also be derived for $\matr{\Theta}_{gnn}$.

With $\Upsilon / \sqrt{m} \leq 1$ and $\Lambda^{(j)}\leq \beta_{h}$ and a similar procedure as in Lemma \ref{lemma_after_GD_model_results_variables} and Lemma \ref{lemma_after_GD_weight_matrices_bounds}, we have
\begin{displaymath}
\begin{split}
    &\norm{\matr{\Theta}^{(j+1)} - \matr{\Theta}^{(j)}}_{F} \leq \eta \frac{\Upsilon^{(j)}}{\sqrt{m}},
    \quad \norm{\matr{\Theta}^{(j)}}_{F} \leq 2\beta_{L}\sqrt{m}  \\
    & \norm{\vect{h}^{(j+1)} - \vect{h}^{(j)}}_{2} \leq \eta\frac{2\zeta \beta_{h}}{\sqrt{m}}(2\zeta \beta_{L})^{L}\Upsilon^{(j)},
    \quad \norm{\vect{h}^{(j)}}_{2} \leq 2 \beta_{h}, \\
    & \norm{\matr{\Gamma}^{(j+1)} - \matr{\Gamma}^{(j)}}_{F} \leq 2\eta\zeta^{2} \beta_{h}(2\zeta \beta_{L})^{L}\Upsilon^{(j)}, \quad \norm{\matr{\Gamma}^{(j)}}_{2} \leq \zeta 
\end{split}
\end{displaymath}
With Lemma G.1 from \cite{skip_kernel_du2019gradient}, for $\matr{\Theta}\in \{\matr{\Theta}_{gnn}, \matr{\Theta}_{1}, \dots, \matr{\Theta}_{L}\}$,  
\begin{displaymath}
\begin{split}
    \norm{\nabla f({\matr{\Theta}}^{(j)}) - \nabla f({\matr{\Theta}}^{(j)}, s)}_{F} \leq 
    \frac{4\zeta}{\sqrt{m}} \eta\Upsilon^{(j)}\beta_{h}L
    (2\zeta \beta_{L})^{2L}.
\end{split}
\end{displaymath}
Combining with $\norm{\nabla\mathcal{L}({\Theta'}^{(j)})}_{F}$, we have
\begin{displaymath}
\begin{split}
    \abs{\matr{V}^{(j)}(\widetilde{\matr{X}})} \leq \eta^{2}\frac{4T}{m} (L+2)^{2} \beta_{h}^{3} \cdot\norm{\vect{F}^{(j)}_{T} - \vect{Y}_{T}}_{2}^{2} (2\zeta \beta_{L})^{4L}.
\end{split}
\end{displaymath}
Since this inequality holds for an arbitrary $\widetilde{\matr{X}}\in \{\widetilde{\matr{X}}_{\tau}\}_{\tau\in [T]}$ and $\norm{\vect{F}^{(0)}_{T} - \vect{Y}_{T}}_{2}=\mathcal{O}(\sqrt{T})$, given network width $m$, we finally have 
\begin{displaymath}
\begin{split}
    \norm{\matr{V}^{(j)}} \leq \frac{1}{4}\eta \beta_{F} \norm{\vect{F}^{(j)}_{T} - \vect{Y}_{T}}_{2}^{2}.
\end{split}
\end{displaymath}
with the choice of learning rate $\eta \leq \mathcal{O}(T^{-1}L^{-1}\beta_{h}^{-2}(2\zeta \beta_{L})^{-2L})$.
$\blacksquare$


\textbf{Proof of Lemma \ref{lemma_after_GD_next_iteration_output}.}
We prove this lemma following an analogous approach as Lemma B.7 in \cite{skip_kernel_du2019gradient}.
By the model definition and substituting $\Upsilon^{(j)} /\sqrt{m}$ with $m^{-\frac{1}{2}} 2\sqrt{T} \norm{\vect{F}^{(j)}_{T}- \vect{Y}_{T}}_{2} \cdot \zeta^{L} (2\beta_{L})^{L} \beta_{h}$ as the upper bound based on Lemma \ref{lemma_after_GD_weight_matrices_bounds}, with $\Lambda^{(j)} \leq \beta_{h}$, we have
\begin{displaymath}
\begin{split}
    & \norm{\vect{F}^{(j)}_{T} - \vect{F}^{(j+1)}_{T}}_{2}^{2} = 
    \frac{1}{m}\sum_{\tau=1}^{T} \big( (\vect{h}_{L-1, \tau}^{(j+1)})^{\intercal}\matr{\Theta}_{L}^{(j+1)} -
    (\vect{h}_{L-1, \tau}^{(j)})^{\intercal}\matr{\Theta}_{L}^{(j)}\big)^{2} \\
    & \leq \frac{2}{m}\bigg(\norm{\matr{\Theta}_{L}^{(j+1)} - \matr{\Theta}_{L}^{(j)}}_{2}^{2} 
    \sum_{\tau=1}^{T} \norm{\vect{h}_{L-1, \tau}^{(j+1)}}_{2}^{2} + 
    \norm{\matr{\Theta}_{L}^{(j)}}_{2}^{2} \sum_{\tau=1}^{T} \norm{\vect{h}_{L-1, \tau}^{(j+1)} - \vect{h}_{L-1, \tau}^{(j)}}_{2}^{2}\bigg) \\
    & \leq \frac{2}{m}\bigg(
    \frac{T}{m}\eta^{2} (2\beta_{h})^{4} \norm{\vect{F}^{(j)}_{T} - \vect{Y}_{T}}_{2}^{2}
    + T(2\beta_{3})^{2} (\eta\frac{2\zeta \beta_{h}}{\sqrt{m}}(2\zeta \beta_{L})^{L}\Upsilon^{(j)})^{2} 
    \bigg) \\
    & \leq \frac{1}{4}\eta \beta_{F} \norm{\vect{F}^{(j)}_{T} - \vect{Y}_{T}}_{2}^{2}
\end{split}
\end{displaymath}
where the last inequality is due to sufficiently large $m$ and the choice of learning rate $\eta$.  $\blacksquare$

\end{document}